\documentclass{article}

\usepackage{microtype}
\usepackage{graphicx}
\usepackage{booktabs}
\usepackage{xcolor}
\usepackage{hyperref}

\usepackage{amsmath}
\usepackage{amsthm}
\usepackage{amssymb}
\usepackage{bbm} 
\usepackage{bm}
\usepackage{verbatim}
\usepackage{float}
\usepackage{soul}
\usepackage{enumitem}
\usepackage{mathtools}
\usepackage{hhline}
\usepackage[title]{appendix}
\usepackage[nameinlink]{cleveref} %
\usepackage[font=small,labelfont=bf,
   justification=justified,
   format=plain]{caption}
\usepackage{subcaption}
\usepackage{tabularx}
\usepackage{tikz}
\usepackage{wrapfig,booktabs}
\usepackage{xspace}
\usepackage{cancel}
\usepackage{sidecap}

\newcommand{\dgps}{\textsc{dgp}s\xspace}
\newcommand{\dgp}{\textsc{dgp}\xspace}
\newcommand{\gps}{\textsc{gp}s\xspace}
\newcommand{\gp}{\textsc{gp}\xspace}

\newcommand{\elbo}{\textsc{elbo}\xspace}

\crefname{appsec}{appendix}{appendices}
\Crefname{appsec}{Appendix}{Appendices}

\usetikzlibrary{shapes.geometric, arrows, bayesnet, calc, positioning}

\DeclareMathOperator{\xbf}{\mathbf{x}}
\DeclareMathOperator{\Xbf}{\mathbf{X}}
\DeclareMathOperator{\ybf}{\mathbf{y}}

\DeclareMathOperator{\fbf}{\mathbf{f}}

\DeclareMathOperator{\ubf}{\mathbf{u}}

\DeclareMathOperator{\zbf}{\mathbf{z}}
\DeclareMathOperator{\Zbf}{\mathbf{Z}}

\DeclareMathOperator{\ku}{\mathbf{k_{u}}}

\DeclareMathOperator{\Kuu}{\mathbf{K_{uu}}}

\DeclareMathOperator{\dfbf}{\text{d}\mathbf{f}}

\newcommand{\dee}{\,\textrm{d}}

\newcommand{\calO}{\mathcal{O}}
\newcommand{\calL}{\mathcal{L}}

\newcommand{\closer}[3]{{\kern-#1ex{#2}\kern-#3ex}}

\mathchardef\mhyphen="2D

\DeclareMathOperator{\E}{\mathbb{E}}

\usepackage{float}

\newcommand\defines{\stackrel{\mathclap{\normalfont\mbox{\tiny def}}}{=}}

\newcommand{\dsvi}{\textsc{dsvi}\xspace}
\newcommand{\svi}{\textsc{svi}\xspace}
\newcommand{\rkhs}{\textsc{rkhs}\xspace}
\newcommand{\vff}{\textsc{vff}\xspace}

\usetikzlibrary{shapes.geometric, arrows, bayesnet, calc, positioning}

\tikzstyle{arrow} = [thick,->,>=stealth]

\tikzstyle{sparse} = [rectangle, rounded corners, minimum width=2cm, minimum height=1cm,text centered, text width=3.5cm, draw=black, fill=red!50]

\tikzstyle{variational} = [rectangle, rounded corners, minimum width=3cm, minimum height=1cm,text centered, text width=3.5cm, draw=black, fill=blue!50]

\tikzstyle{deep} = [rectangle, rounded corners, minimum width=3cm, minimum height=1cm,text centered, text width=3.5cm, draw=black, fill=violet!50]

\tikzstyle{input neuron}=[neuron, fill=green!15]
\tikzstyle{output neuron}=[neuron, fill=red!15]
\tikzstyle{hidden neuron}=[neuron, fill=blue!15]

\usepackage[accepted]{include/icml2020}

\allowdisplaybreaks

\icmltitlerunning{Inter-domain Deep Gaussian Processes}

\begin{document}

\twocolumn[

\icmltitle{
    Inter-domain Deep Gaussian Processes
}

\begin{icmlauthorlist}
    \icmlauthor{Tim G. J. Rudner}{oxfordcs}
    \icmlauthor{Dino Sejdinovic}{oxfordstats}
    \icmlauthor{Yarin Gal}{oxfordcs}
\end{icmlauthorlist}

\icmlaffiliation{oxfordcs}{
    Department of Computer Science, University of Oxford, Oxford, United Kingdom
}
\icmlaffiliation{oxfordstats}{
    Department of Statistics, University of Oxford, Oxford, United Kingdom
}

\icmlcorrespondingauthor{Tim G. J. Rudner}{tim.rudner@cs.ox.ac.uk}

\icmlkeywords{
    Machine Learning, Gaussian Processes, Deep Gaussian Processes, Variational Inference
}

\vskip 0.3in
]

\printAffiliationsAndNotice{}

\begin{abstract}
    Inter-domain Gaussian processes (\gps) allow for high flexibility and low computational cost when performing approximate inference in \gp models. They are particularly suitable for modeling data exhibiting global structure but are limited to stationary covariance functions and thus fail to model non-stationary data effectively. We propose \textit{Inter-domain Deep Gaussian Processes}, an extension of inter-domain shallow \gps that combines the advantages of inter-domain and deep Gaussian processes (\dgps), and demonstrate how to leverage existing approximate inference methods to perform simple and scalable approximate inference using inter-domain features in \dgps. We assess the performance of our method on a range of regression tasks and demonstrate that it outperforms inter-domain shallow \gps and conventional \dgps on challenging large-scale real-world datasets exhibiting \textit{both} global structure \textit{as well as} a high-degree of non-stationarity.
\end{abstract}


\section{Introduction}

Gaussian processes (\gps) are a powerful tool for function approximation. They are Bayesian non-parametric models and as such they are flexible, robust to overfitting, and provide well-calibrated predictive uncertainty estimates \citep{Rasmussen2005,Bui2016}.
Deep Gaussian processes (\dgps) are layer-wise compositions of \gps designed to model a larger class of functions than shallow \gps.

To scale \gp and \dgp models to large datasets, a wide array of approximate inference methods has been developed, with inducing points-based variational inference being the most widely used \citep{Snelson2005,Titsias2009,Wilson2015kernel}.
However, conventional inducing points-based inference for \gps relies on \textit{point} evaluations and thus, by construction, creates \textit{local} approximations to the target function.
As a result, the approximate posterior predictive distribution may fail to capture complex \textit{global} structure in the data, severely limiting the usefulness and computational efficiency of local inducing points-based approximations.

Inter-domain \gps were  designed to overcome this limitation.
In order to capture global structure in the underlying data-generating process, inter-domain \gps define inducing variables as projections of the target function over the entire input space and not as mere point evaluations \citep{Lazaro-Gredilla-IDGP,Rahimi2007,Gal2015}.
The resulting posterior predictive distribution is able to represent complex data with global structure with higher accuracy as local approximations but at the same computational cost.
Unfortunately, inter-domain projections most suitable for capturing global structure (e.g., spectral transforms) are limited by the fact that they can only be used with stationary covariance functions, making them ill-suited for modeling non-stationary data and limiting their usefulness in practice.

\begin{figure*}[t!]
\centering
\begin{subfigure}{0.31\linewidth}
    \centering
    \includegraphics[width=\columnwidth]{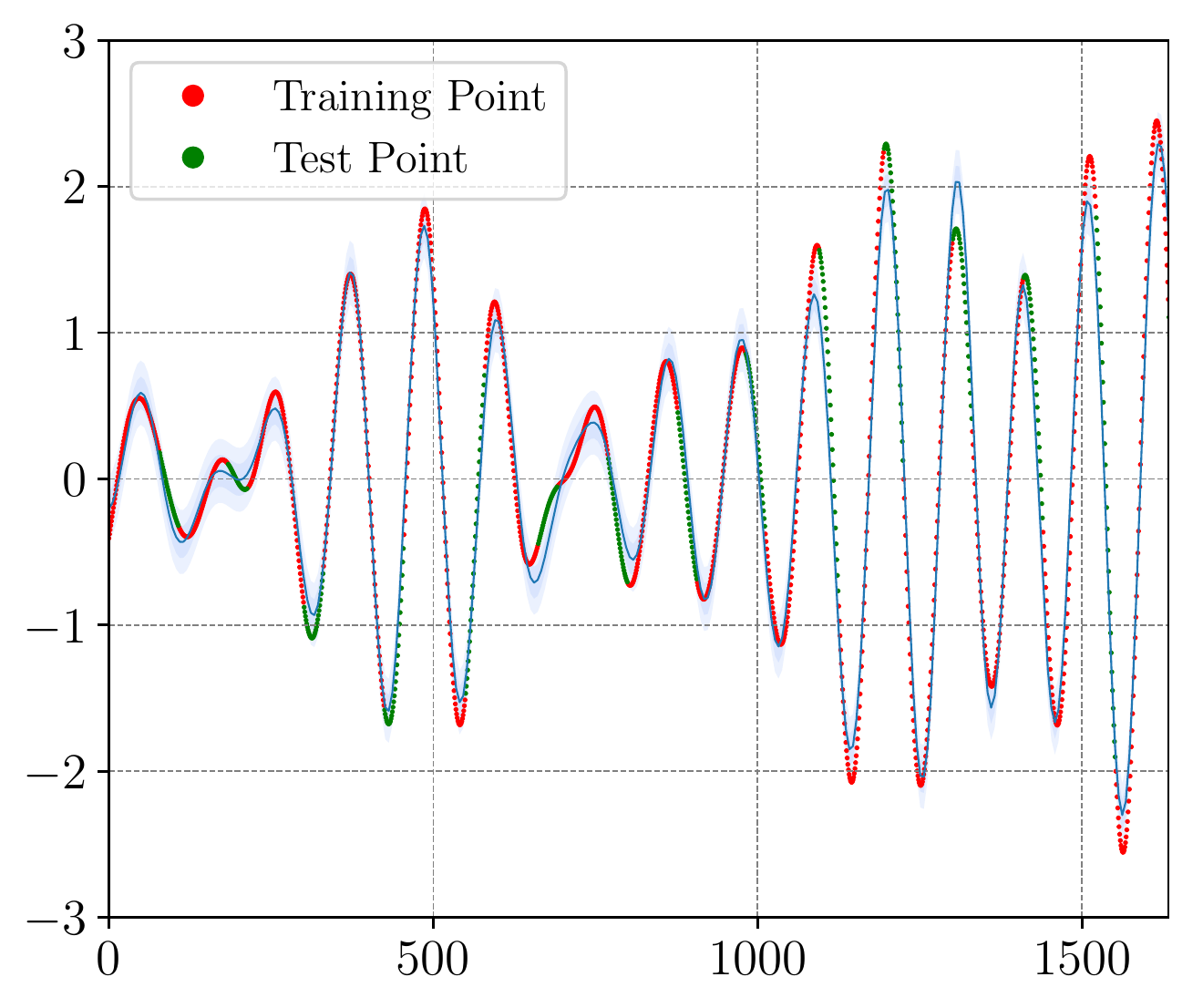}
    \caption{
        Deep \gp with global approximations.
    }
    \label{fig:audio-pred-a}
\end{subfigure}
~\hspace{0.01cm}
\begin{subfigure}{0.31\linewidth}
    \centering
    \includegraphics[width=\columnwidth]{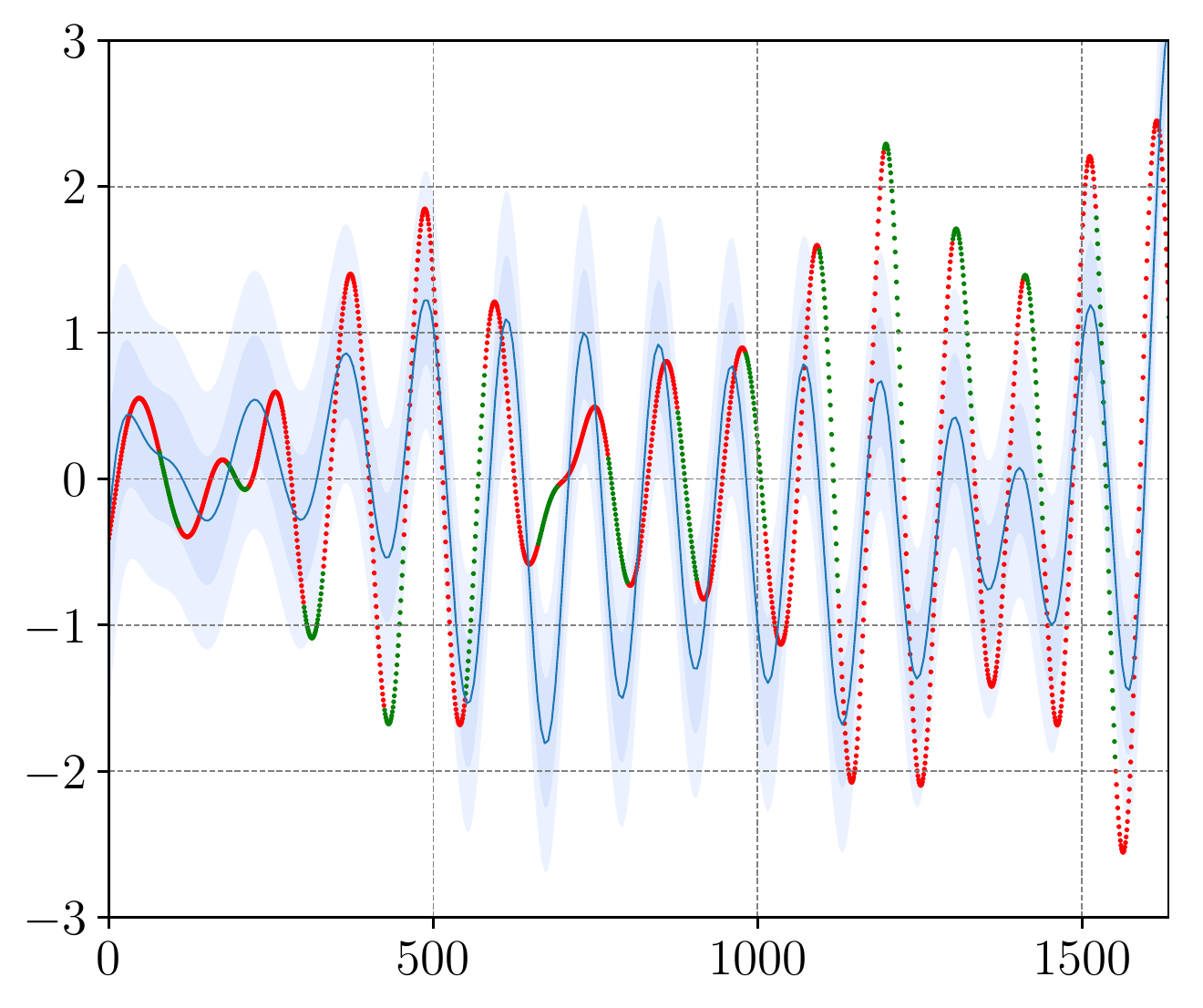}
    \caption{
        Shallow \gp with global approximations.\hspace*{-10pt}
    }
    \label{fig:audio-pred-b}
\end{subfigure}
~\hspace{0.01cm}
\begin{subfigure}{0.31\linewidth}
    \centering
    \includegraphics[width=\columnwidth]{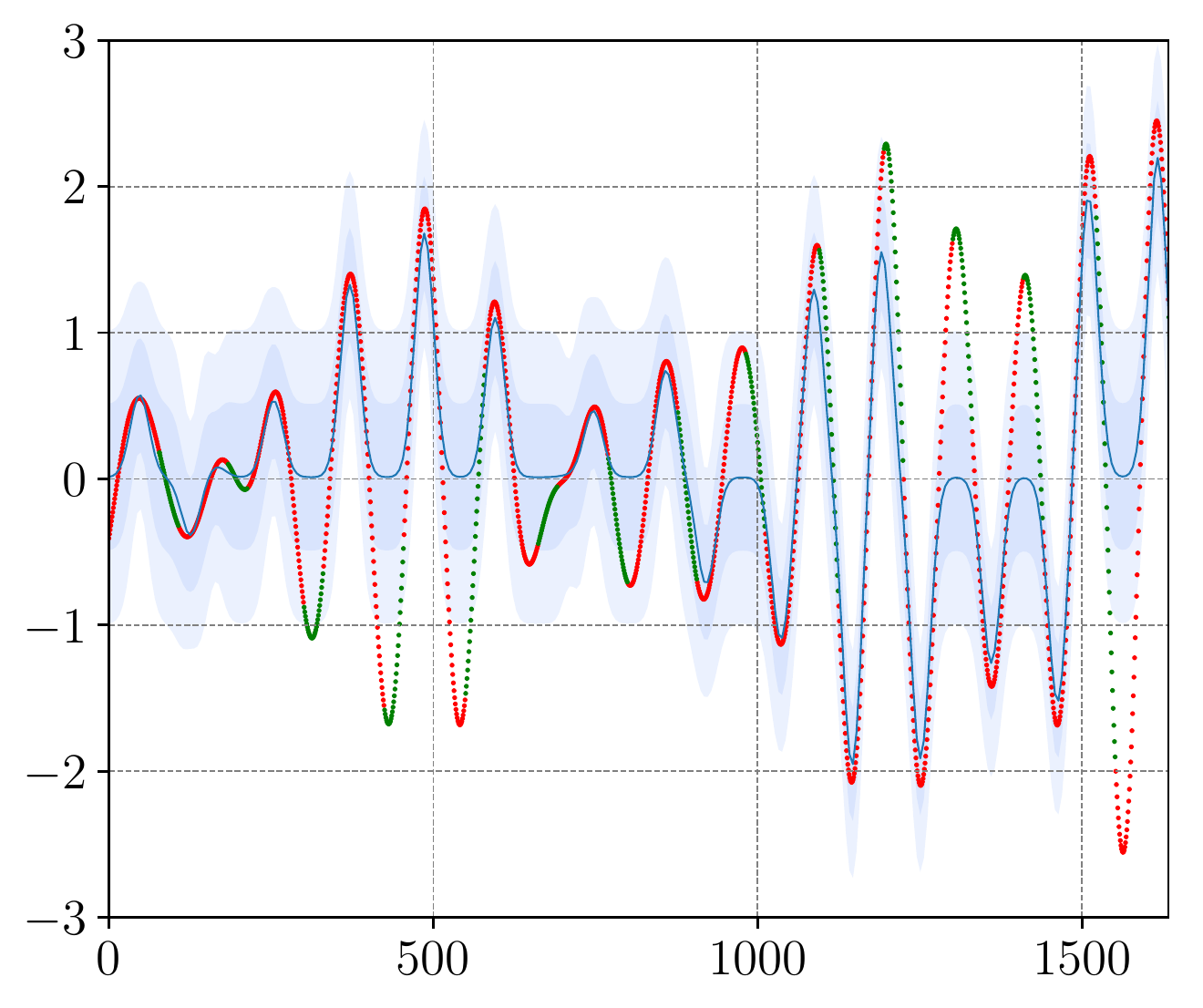}
    \caption{
        Deep \gp with local approximations.
    }
    \label{fig:audio-pred-c}
    \end{subfigure}
\caption{
    Approximate posterior predictive distributions for shallow and deep \gp models obtained from 20 inducing points.
    The blue lines denote the posterior predictive means of the models, respectively.
    Each shade of blue corresponds to one posterior standard deviation.
}
\label{fig:audio-pred}
\end{figure*}

We propose \textit{Inter-domain Deep Gaussian Processes} to overcome this limitation while retaining the benefits of inter-domain methods.\footnote{For source code and additional results, see \url{https://bit.ly/inter-domain-dgps}.}
Specifically, we define an augmented \dgp model, in which we replace local inducing variables by reproducing kernel Hilbert space (\rkhs) Fourier features, and exploit the compositional structure of the variational distribution in doubly stochastic variational inference (\dsvi) for \dgps.
This way, we achieve simple and scalable approximate inference while efficiently capturing global structure in the underlying data-generating process.
The resulting inter-domain \dgp is composed of a composition of inter-domain \gps, which makes it possible to efficiently model complex, non-stationary data despite each inter-domain \gp in the hierarchy being restricted to stationary covariance functions.

We establish that our method performs well on several complex real-world datasets exhibiting global structure and non-stationarity and demonstrate that inter-domain \dgps are more computationally efficient than \dgps with local approximations when modeling data with global structure.
\Cref{fig:audio-pred} shows approximate posterior predictive distributions of an inter-domain deep \gp (\ref{fig:audio-pred-a}), an inter-domain shallow \gp (\ref{fig:audio-pred-b}), and a deep \gp based on local approximations (\ref{fig:audio-pred-c}) on a dataset with global structure.

To summarize, our main contributions are as follows:
\begin{enumerate}
    \item We propose \textit{Inter-domain Deep Gaussian Processes} and use \rkhs Fourier features to incorporate global structure into the \dgp posterior predictive distribution;
    \item We present a simple approach for performing approximate inference in inter-domain \dgps by exploiting the compositional structure of the variational distribution in \dsvi;
    \item We show that inter-domain \dgps significantly outperform both inter-domain shallow \gps and state-of-the-art local approximate inference methods for \dgps on complex real-world datasets with global structure;
    \item We demonstrate that inter-domain \dgps are more computationally efficient than local approximate inference methods for \dgps when trained on data exhibiting global structure.
\end{enumerate}


\section{Background}

We begin by reviewing \dgps and inter-domain \gps.
We will draw on this exposition in subsequent sections.


\subsection{Deep Gaussian Processes}

\dgps are layer-wise compositions of \gps in which the output of a previous layer is used as the input to the next layer.
Similar to deep neural networks, the hidden layers of a \dgp learn representations of the input data, but unlike neural networks, they allow for uncertainty to be propagated through the function compositions.
This way, \dgps define probabilistic predictive distributions over the target variables and---unlike for shallow \gps---any finite collection of random variables distributed according to a \dgp posterior predictive distribution does not need to be jointly Gaussian, allowing \dgp models to represent a larger class of distributions over functions than shallow \gps.

Consider a set of $N$ noisy target observations $\ybf \in \mathbb{R}^N$ at corresponding input points \mbox{$\Xbf = [\xbf_1, ..., \xbf_N]^\top \in \mathbb{R}^{N\times D}$}.
A \dgp is defined by the composition
\begin{equation}
\begin{aligned}
\label{eq:dgpcomposition}
    \ybf = \fbf^{(L)} + \bm{\epsilon} \defines f^{(L)}(f^{(L-1)}(...f^{(1)}(\Xbf))...) + \bm{\epsilon},
\end{aligned}
\end{equation}
where $L$ is the number of layers, $\bm{\epsilon} \sim \mathcal{N}({\mathbf{0}, \sigma^{2} \mathbf{I})}$, and $\fbf^{(\ell)} = f^{(\ell)}(\fbf^{(\ell - 1)})$ in the composition denotes the $\ell$th-layer \gp, $f^{(\ell)}(\cdot)$, evaluated at $\fbf^{(\ell - 1)}$.
We follow previous work and absorb the noise between layers, which is assumed to be i.i.d. Gaussian, into the kernel so that $k_{\text{noisy}} (\xbf_i, \xbf_j) = k(\xbf_i, \xbf_j) + \sigma^{2}_{(\ell)} \delta_{ij}$, where $\delta_{ij}$ is the Kronecker delta and $\sigma^{2}_{(\ell)}$ is the noise variance between layers \citep{Salimbeni2017}.

A \dgp with likelihood $p(\ybf | \fbf^{(L)})$ has the joint distribution
\begin{align*}
    p(\ybf, \{\fbf^{(\ell)} \}_{\ell=1}^L ) = &\prod_{n=1}^N p(\ybf_n | \fbf_n^{(L)}) \prod_{\ell = 1}^{L} p(\fbf^{(\ell)} | \fbf^{(\ell - 1)} ),
\end{align*}
with $\fbf^0 \defines \Xbf$.
Unlike shallow \gps, exact inference in \dgps is not analytically tractable due to the nonlinear transformations at every layer of the composition in~\Cref{eq:dgpcomposition}.

To make posterior inference tractable, a number of approximate inference techniques for \dgps have been developed with the aim of improving performance, scalability, stability, and ease of optimization \citep{Dai2015, Hensman2014, Bui2016, Salimbeni2017, Cutajar2017, Mattos2015, Havasi2018, Salimbeni2019}.


\subsection{Inter-domain Gaussian Processes}
\label{sec:idgp}

Inter-domain \gps are centered around the idea of finding a possibly more compact representative set of input features in a domain different from the input data domain.
This way, it is possible to incorporate prior knowledge about relevant characteristics of data---such as the presence of global structure---into the inducing variables.

Consider a real-valued \gp $f(\xbf)$ with $\xbf \in \mathbb{R}^D$ and some deterministic function $g(\xbf, \Zbf)$, with $M$ inducing points \mbox{$\Zbf \in \mathbb{R}^{M \times H}$}.
We define the following transformation:
\begin{align}
\label{eq:inter_domain_transformation}
    u(\mathbf{Z})
    \defines
    \int_{\mathbb{R}^{D}} f(\mathbf{x}) g(\mathbf{x}, \mathbf{Z}) \dee \mathbf{x}.
\end{align}
Since $u(\Zbf)$ is obtained through an affine transformation of $f(\xbf)$, $u(\Zbf)$ is also a \gp, but may lie in a different domain than $f(\xbf)$ \citep{Lazaro-Gredilla-IDGP}.
Inter-domain \gps arise when $f(\xbf)$ and $u(\Zbf)$ are considered \textit{jointly} as a single, augmented \gp, as is the case for local inducing points-based approximate inference.
The feature extraction function $g(\xbf, \Zbf)$ used in the integral then defines the transformed domain in which the inducing dataset lies.
The inducing variables obtained this way can be seen as projections of the target function $f(\xbf)$ on the feature extraction function over the entire input space \citep{Lazaro-Gredilla-IDGP}.
As such, each of the inducing variables is constructed to contain information about the structure of $f(\xbf)$ everywhere in the input space, making them more informative of the stochastic process than local approximations \citep{Hensman2016, Lazaro-Gredilla-IDGP}.

In general, the usefulness of inducing variables mostly relies on their covariance with the remainder of the process, which, for inducing points-based approximate inference, is encoded in the vector-valued function
\begin{align*}
    \ku(\xbf) = [ k(\zbf_1, \xbf), k(\zbf_2, \xbf), ..., k(\zbf_M, \xbf)].
\end{align*}
The matrix $\Kuu \defines k(\Zbf, \Zbf)$ and the vector-valued function $\ku(\xbf)$ are central to inducing points-based approximate inference for \gps where they are used to construct an approximate posterior distribution.

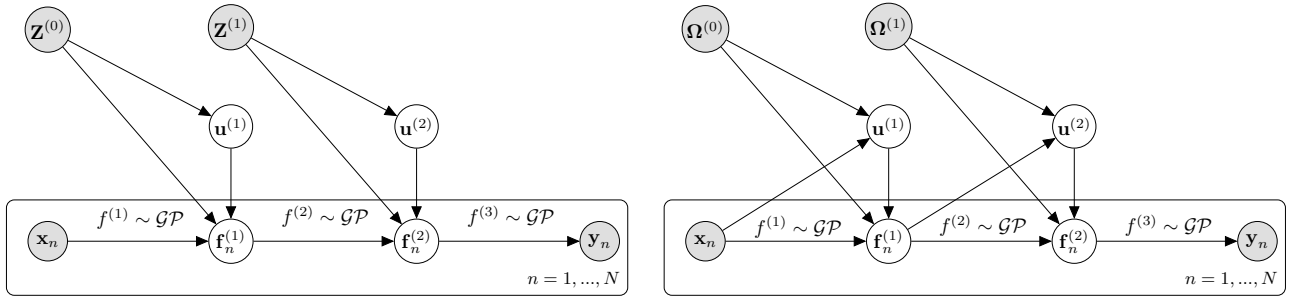
\begin{figure*}[th!]
    \centering
    \begin{subfigure}[t]{0.49\textwidth}
    \resizebox{1\textwidth}{!}{
    \begin{tikzpicture}[scale = 0.1]
    \node[obs] (x) {$\xbf_n$} ; %
    \node[latent, right=of x, xshift=1.5cm] (h1) {$\fbf_n^{(1)}$} ; %
    \node[latent, right=of h1, xshift=1.5cm] (h2) {$\fbf_n^{(2)}$} ; %
    \node[obs, right=of h2, xshift=1.5cm] (y) {$\ybf_n$} ; %
    \node[const, above right=of x, xshift=-0.1cm, yshift=-0.75cm] (f1) {$f^{(1)} \sim \mathcal{GP}$} ; %
    \node[const, above right=of h1, xshift=-0.1cm, yshift=-0.75cm] (f2) {$f^{(2)} \sim \mathcal{GP}$} ; %
    \node[const, above right=of h2, xshift=-0.1cm, yshift=-0.75cm] (f3) {$f^{(3)} \sim \mathcal{GP}$} ; %
    \node[latent, above=of h1, yshift=0.25cm] (u1) {$\ubf^{(1)}$} ; %
    \node[latent, above=of h2, yshift=0.25cm] (u2) {$\ubf^{(2)}$} ; %
    \node[obs, above=of x, yshift=2cm] (z0) {$\Zbf^{(0)}$} ; %
    \node[obs, above=of h1, yshift=2cm] (z1) {$\Zbf^{(1)}$} ; %
    \plate[inner sep=0.25cm, xshift=-0.12cm, yshift=0.12cm] {plate1} {(x) (y)} {$n=1,...,N$}; %
    \edge {x} {h1} ; %
    \edge {h1} {h2} ; %
    \edge {h2} {y} ; %
    \edge {z0} {u1} ; %
    \edge {z0} {h1} ; %
    \edge {z1} {u2} ; %
    \edge {z1} {h2} ; %
    \edge {u1} {h1} ; %
    \edge {u2} {h2} ; %
   \end{tikzpicture}
   }
   \caption{Graphical model representation of a \dgp model with local inducing inputs, inducing variables, and two hidden layers, $\fbf^{(1)}$ and $\fbf^{(2)}$, for $n=1,...,N$.} \label{fig:graphdsvdgp}
   \end{subfigure}
   \hfill
   \begin{subfigure}[t]{0.49\textwidth}
   \resizebox{1\textwidth}{!}{
    \begin{tikzpicture}[scale = 0.1]
    \node[obs] (x) {$\xbf_n$} ; %
    \node[latent, right=of x, xshift=1.5cm] (h1) {$\fbf_n^{(1)}$} ; %
    \node[latent, right=of h1, xshift=1.5cm] (h2) {$\fbf_n^{(2)}$} ; %
    \node[obs, right=of h2, xshift=1.5cm] (y) {$\ybf_n$} ; %
    \node[const, above right=of x, xshift=-0.1cm, yshift=-0.9cm] (f1) {$f^{(1)} \sim \mathcal{GP}$} ; %
    \node[const, above right=of h1, xshift=-0.1cm, yshift=-0.9cm] (f2) {$f^{(2)} \sim \mathcal{GP}$} ; %
    \node[const, above right=of h2, xshift=-0.1cm, yshift=-0.9cm] (f3) {$f^{(3)} \sim \mathcal{GP}$} ; %
    \node[latent, above=of h1, yshift=0.25cm] (u1) {$\ubf^{(1)}$} ; %
    \node[latent, above=of h2, yshift=0.25cm] (u2) {$\ubf^{(2)}$} ; %
    \node[obs, above=of x, yshift=2cm] (omega0) {$\bm{\Omega}^{(0)}$} ; %
    \node[obs, above=of h1, yshift=2cm] (omega1) {$\bm{\Omega}^{(1)}$} ; %
    \plate[inner sep=0.25cm, xshift=-0.12cm, yshift=0.12cm] {plate1} {(x) (y)} {$n=1,...,N$}; %
    \edge {x} {h1} ; %
    \edge {h1} {h2} ; %
    \edge {h2} {y} ; %
    \edge {omega0} {u1} ; %
    \edge {omega1} {u2} ; %
    \edge {u1} {h1} ; %
    \edge {u2} {h2} ; %
    \edge {omega0} {h1}
    \edge {omega1} {h2}
    \edge {x} {u1}
    \edge {h1} {u2}
  \end{tikzpicture}
  }
  \caption{Graphical model representation of an inter-domain \dgp model with inducing frequencies, \rkhs Fourier feature inducing variables, and two hidden layers, $\fbf^{(1)}$ and $\fbf^{(2)}$, for $n=1,...,N$.} \label{fig:graphdsvffdgp}
\end{subfigure}
\caption{Graphical model representations of local inducing-points \dgps (\Cref{fig:graphdsvdgp}) and inter-domain \dgps (\Cref{fig:graphdsvffdgp}). Greyed-out nodes denote observed data and non-greyed out nodes denote unobserved data.}
\end{figure*}


\section{Inter-domain Deep Gaussian Processes}

In this section, we will introduce inter-domain \dgps.
First, we will present a general inter-domain \dgp framework.
Next, we will explain why constructing inter-domain deep \gps is more challenging than constructing inter-domain shallow \gps and how we can leverage the compositional structure of the layer-wise approximate posterior predictive distributions in doubly stochastic variational inference \citep{Salimbeni2017} to obtain simple and scalable inter-domain \dgps.
Finally, we will draw on prior work \citep{Hensman2016} to explicitly incorporate global structure into the inter-domain transformation.


\subsection{The Augmented Inter-domain Deep Gaussian Process Model}

In inducing points-based approximate inference, the \gp model is augmented by a set of inducing variables, $u(\Zbf)$.
Unlike conventional inducing points-based approximations, inter-domain approaches do not constrain inducing points to lie in the same domain as the input data.

To distinguish between inducing points that lie in the same domain as the input data and inter-domain inducing points, we diverge from the notation in the previous section and henceforth denote inter-domain inducing points across \dgp layers as \mbox{$\{ \bm{\Omega}^{(\ell)} \}_{\ell=0}^{L-1}$} with corresponding inducing variables \mbox{$\ubf^{(\ell)} \defines u(\bm{\Omega}^{(\ell-1)})$} for \mbox{$\ell=1, ..., L$}, where $L$ is the number of \dgp layers.
Inter-domain \dgps arise when considering $f^{(\ell)}(\fbf^{(\ell-1)})$ and $u(\bm{\Omega}^{(\ell-1)})$ jointly as a single, extended \gp.
We can then express the extended \dgp joint distribution as
\begin{align}
\label{eq:dgp_joint}
    p(\ybf, \{ \fbf^{(\ell)}, \ubf^{(\ell)} \}_{\ell = 1}^ {L} ) = &\prod_{n=1}^N p(\ybf_n | \fbf_n^{(L)})
    \nonumber
    \\
    &~~~
    \cdot \prod_{\ell =1}^{L} p(\fbf^{(\ell)} | \ubf^{(\ell)}; \fbf^{(\ell - 1)}, \bm{\Omega}^{(\ell - 1)})
    \nonumber
    \\
    &~~~~~~
    \cdot p(\ubf^{(\ell)} ; \fbf^{(\ell - 1)}, \bm{\Omega}^{(\ell - 1)}).
\end{align}
Importantly, in contrast to conventional inducing points-based approaches, each $p(\ubf^{(\ell)} ; \fbf^{(\ell - 1)}, \bm{\Omega}^{(\ell - 1)})$ above is a distribution over the (inter-domain) inducing variables and as such contains information about $\fbf^{(\ell - 1)}$ via the inter-domain projection defined in~\Cref{eq:inter_domain_transformation}.
For a graphical representation, see~\Cref{fig:graphdsvffdgp}.

Analogously to inter-domain \gps, the mean and covariance functions at each layer of this extended \dgp accept arguments from both the input and transformed domains and treat them accordingly.
Following prior work, we refer to inter-domain mean and covariance functions as \emph{instances}, which accept different types of arguments \citep{Lazaro-Gredilla-IDGP}.
For each \dgp layer, the transformed-domain instance of the mean function is given by
\begin{align}
    &m^{(\ell)}(\bm{\Omega}^{(\ell - 1)})
    \nonumber
    \\
    &~~~=
    \mathbb{E} [u(\bm{\Omega}^{(\ell-1)})]
    \nonumber
    \\
    &~~~=
    \int_{\mathbb{R}^{D^{(\ell-1)}}} \mathbb{E}[ f^{(\ell)}(\fbf^{(\ell - 1)}) ] \, g(\fbf^{(\ell - 1)}, \bm{\Omega}^{(\ell - 1)}) \dfbf^{(\ell - 1)}
    \nonumber
    \\
    &~~~=
    \int_{\mathbb{R}^{D^{(\ell-1)}}} m^{(\ell)}(\fbf^{(\ell - 1)}) \, g(\fbf^{(\ell - 1)}, \bm{\Omega}^{(\ell - 1)}) \dfbf^{(\ell - 1)}
    \nonumber
    \\
    &~~~\defines
    \mathbf{m^\phi_{u^{\ell}}},
\end{align}
and the transformed-domain instances of the covariance function are
\begin{align}
    &k^{(\ell)}(\fbf^{(\ell-1)}, \bm{\Omega}^{(\ell - 1')})
    \nonumber
    \\
    &=
    \mathbb{E} [f^{(\ell)}(\fbf^{(\ell-1)})  u(\bm{\Omega}^{(\ell-1)'})]
    \nonumber
    \\
    &=
    \mathbb{E} \bigg[ f^{(\ell)}(\fbf^{(\ell - 1)})
    \nonumber
    \\
    &\qquad\,\,\,\,
    \cdot \int_{\mathbb{R}^{D^{(\ell-1)}}} f^{(\ell)}(\fbf^{(\ell - 1)'}) \, g(\fbf^{(\ell - 1)'}, \bm{\Omega}^{(\ell - 1)'}) \dfbf^{(\ell - 1)'} \bigg]
    \nonumber
    \\
    &=
    \int_{\mathbb{R}^{D^{(\ell-1)}}} k^{(\ell)}(\fbf^{(\ell - 1)}, \fbf^{(\ell - 1)'}) \, g(\fbf^{(\ell - 1)'}, \bm{\Omega}^{(\ell - 1)'}) \dfbf^{(\ell - 1)'}
    \nonumber
    \\
    &\defines \mathbf{K}^\phi_{\mathbf{f^{\ell} u^{\ell}}}
\end{align}
and
\begin{align}
    &k^{(\ell)}(\bm{\Omega}^{(\ell - 1)}, \bm{\Omega}^{(\ell - 1)'})
    \nonumber
    \\
    &=
    \mathbb{E} [u(\bm{\Omega}^{(\ell - 1)}) u(\bm{\Omega}^{(\ell-1)'})]
    \nonumber
    \\
    &=
    \mathbb{E} \bigg[ \int_{\mathbb{R}^{D^{(\ell-1)}}} f^{(\ell)}(\fbf^{(\ell - 1)}) \, g(\fbf^{(\ell - 1)}, \bm{\Omega}^{(\ell - 1)}) \dfbf^{(\ell - 1)}
    \nonumber
    \\
    &\qquad\,\,\,\,
    \cdot \int_{\mathbb{R}^{D^{(\ell-1)}}} f^{(\ell)}(\fbf^{(\ell - 1)'}) \, g(\fbf^{(\ell - 1)'}, \bm{\Omega}^{(\ell - 1)'}) \dfbf^{(\ell - 1)'} \bigg]
    \nonumber
    \\
    &=
    \int_{\mathbb{R}^{D^{(\ell-1)}}} \int_{\mathbb{R}^{D^{(\ell-1)}}} k^{(\ell)}(\fbf^{(\ell - 1)}, \fbf^{(\ell - 1)'}) \, g(\fbf^{(\ell - 1)}, \bm{\Omega}^{(\ell - 1)})
    \nonumber
    \\
    &\qquad\quad
    \cdot g(\fbf^{(\ell - 1)'}, \bm{\Omega}^{(\ell - 1)'}) \dfbf^{(\ell - 1)} \dfbf^{(\ell - 1)'}
    \nonumber
    \\
    &\defines
    \mathbf{K}^\phi_{\mathbf{u^{\ell} u^{\ell}}},
\end{align}
where $m^{(\ell)}(\cdot)$ and $k^{(\ell)}(\cdot, \cdot)$ denote the $\ell$th layer's mean and covariance function, respectively, $D^{(\ell)}$ is the $\ell$th \dgp layer's output dimension, and we use the superscript $\phi$ to indicate inter-domain instances of the mean and covariance functions.
Mean and covariance functions at each layer are therefore defined both by the values and domains of their arguments.
We can now express the joint distribution in~\Cref{eq:dgp_joint} in terms of the layer-wise covariance functions given above and perform inference across domains.


\subsection{Simple and Scalable Approximate Inference in Inter-domain Deep Gaussian Processes}

Since exact inference in \dgps is intractable, we need approximate inference methods.
Unfortunately, most approximate inference methods for \dgps require computing convolutions between $\mathbf{K}^\phi_{\mathbf{u^{\ell} f}^{\ell}}$ and the distributions of the latent functions, that is,
\begin{align}
    &\int \mathbf{K}^\phi_{\mathbf{u^{\ell} f}^{\ell}} \mathcal{N}(\fbf^{(\ell)} | \mathbf{m_{f^\ell}}, \mathbf{S_{f^\ell}} ) \dfbf^{(\ell)},
\end{align}
where $\mathcal{N}(\fbf^{(\ell)} | \mathbf{m_{f^\ell}}, \mathbf{S_{f^\ell}} )$ represents the variational distribution of layer $\ell$ with mean $\mathbf{m_{f^\ell}}$ and variance $\mathbf{S_{f^\ell}}$ (see, for example, pages 50-51 in \citet{damianou:thesis15} or \citet{Damianou2013}, \citet{Dai2014}, \citet{Bui2016}).
While these convolutions are easy to compute in closed form for conventional inducing points-based approximations where the covariance matrix is computed from the \dgp's input-domain covariance function, they are non-trivial to compute analytically for inter-domain covariance functions \citep{Hensman2016}.

To perform approximate inference in inter-domain \dgps, we exploit the fact that---in contrast to previous inducing points-based variational inference methods for \dgps---the layer-wise marginalization over each $\fbf^{(\ell)}$ in doubly stochastic variational inference (\dsvi) \citep{Salimbeni2017} does not require computing convolutions that explicitly depend on the specific type of cross-covariance function $\mathbf{k}^\phi_{\ubf^{\ell}}(\fbf^{\ell})$.
Instead, the functional form of the posterior predictive distribution $q(\fbf^{(L)})$ and the use of the reparameterization trick make marginalizing out the latent \gp functions across layers straightforward and result in simple, compositional posterior predictive mean and covariance functions at each \dgp layer.
For further details on \dsvi, see~\Cref{appsec:dsvi}.

This property allow us to simply use the inter-domain operators $\mathbf{K}^\phi_{\mathbf{u^{\ell} f^{\ell}}}$ as off-the-shelf replacements for the conventional inducing-point operators $\mathbf{K}_{\mathbf{u^{\ell} f^{\ell}}}$ without having to analytically convolve $\mathbf{K}^\phi_{\mathbf{u^{\ell} f^{\ell}}}$ with the distribution over functions at the $\ell$th layer, yielding the variational distribution
\begin{align*}
\begin{split}
    q(\{ \fbf^{(\ell)} \}^{L}_{\ell = 1}) &= \prod_{\ell = 1}^{L} q(\fbf^{(\ell)} | \bm{\mu}^{(\ell)}, \bm{\Sigma}^{(\ell)}; \fbf^{(\ell - 1)}, \bm{\Omega}^{(\ell - 1)}) \\
    &= \prod_{\ell = 1}^{L} \mathcal{N}(\fbf^{(\ell)} | \widetilde{\mathbf{m}}_{\fbf^{\ell}}, \widetilde{\mathbf{S}}_{\fbf^{\ell}} )
\end{split}
\end{align*}
with
\begin{align}
\label{eq:dgp_posterior_Predictive_layer}
\begin{split}
    \widetilde{\mathbf{m}}_{\fbf^{\ell}} &\defines \mathbf{m}^\phi_{\fbf^{\ell}} - \mathbf{K^\phi_{f^{\ell} u^{\ell}}} \mathbf{K}^{\phi^{-1}}_{\mathbf{u^{\ell} u^{\ell}}} (\bm{\mu}^{(\ell)}- \mathbf{m}^\phi_{\ubf^{\ell}} ), \\
    \widetilde{\mathbf{S}}_{\fbf^{\ell}} &\defines \mathbf{K}^\phi_{\mathbf{f^{\ell} f^{\ell}}} - \mathbf{K^\phi_{f^{\ell} u^{\ell}}}\mathbf{K}^{\phi^{-1}}_{\mathbf{u^{\ell} u^{\ell}}} (\mathbf{K}^\phi_{\mathbf{u^{\ell} u^{\ell}}} - \bm{\Sigma}^{(\ell)}) \mathbf{K}^{\phi^{-1}}_{\mathbf{u^{\ell} u^{\ell}}}\mathbf{K^\phi_{u^{\ell} f^{\ell}}},
\end{split}
\end{align}
where \mbox{$ \mathbf{m_{\fbf^{\ell}}} \defines  m(\fbf^{(\ell-1)})$} and \mbox{$ \mathbf{m_{\ubf^{\ell}}} \defines  m(\bm{\Omega}^{(\ell-1)})$}, and $\bm{\mu}^{(\ell)}$ and $\bm{\Sigma}^{(\ell)}$ are variational parameters.
Since \dsvi uses the reparameterization trick to sample functions at each layer, the inter-domain operators can be used directly to compute the posterior mean and variance for each layer, which allows for simple and scalable approximate inference in inter-domain \dgps.


\subsection{RKHS Fourier Features for Approximate Inference in Gaussian Processes}
\label{sec:rkhsfourierfeatures}

In the previous section, we showed how to perform approximate inference in inter-domain \dgps with any inter-domain operators $\mathbf{k}^\phi_{\mathbf{u}}(x)$ and $\mathbf{K}^\phi_{\mathbf{uu}}$.
Next, we will introduce \rkhs Fourier features \citep{Hensman2016}---an inter-domain approach able to capture global structure in data---and show how to incorporate them into inter-domain \dgps.

\rkhs Fourier features use \rkhs theory to construct inter-domain alternatives to the covariance matrices $\Kuu$ and $\ku(x)$ used in conventional inducing points-based approximate inference methods.
They are constructed by projecting the target function $f$ onto the truncated Fourier basis
\begin{gather}
\begin{aligned}
\label{eq:fourierbasis}
    \bm{\phi}(x)
    \defines &[1, \cos(\omega_1(x - a)), ..., \cos(\omega_M (x - a)), \\
    &\,\, \sin(\omega_1 (x - a)), ..., \sin(\omega_M (x - a))]^\top,
\end{aligned}
\end{gather}
where $x$ is a single, one-dimensional input, and \mbox{$[\omega_1, ..., \omega_M]$} denote inducing \textit{frequencies} defined by \mbox{$\omega_{m} = \frac{2 \pi m}{b-a}$} for some interval $[a,b]$.
From this truncated Fourier basis, we can construct inducing variables as inter-domain projections by defining $u_m \defines \mathcal{P}_{\bm{\phi}_m}(f)$, which can be shown to yield transformed-domain instances of the covariance function given by
\begin{align*}
    \text{cov}(u_m, f(x)) = \bm{\phi}_m(x), ~~~ \text{cov}(u_m, u_{m'}) = \langle \bm{\phi}_{m}, \bm{\phi}_{m'} \rangle_\mathcal{H},
\end{align*}
for both of which there are closed-form expressions if the \gp prior covariance function is given by a half-integer member of the Mat\'ern family of kernels \citep{durrande-periodicities16}.
For further details, see \citet{Hensman2016}.
The resulting inter-domain operators
\begin{align*}
    \mathbf{k}^\phi_{\mathbf{u}}(x) = \bm{\phi}_m (x), \qquad \mathbf{K}^\phi_{\mathbf{uu}} = \langle \bm{\phi}_{m}, \bm{\phi}_{m'} \rangle_\mathcal{H},
\end{align*}
represent inter-domain alternatives to the $\ku(x)$ and $\Kuu$ operators used in local inducing points-based approximations.
By constructing linear combinations of the values of the data-generating process as projections instead of simple function evaluations, the resulting inducing variables become more informative of the underlying process and have more capacity to represent complex functions \citep{Hensman2016,Lazaro-Gredilla-IDGP}.
Analogous to the way in which local inducing points-based approaches approximate the \dgp posterior distribution through kernel functions, \rkhs Fourier features approximate the posterior through sinusoids \citep{Hensman2016}.
The structure imposed by the frequency domain makes \rkhs Fourier features particularly well-suited to capture global structure in data.
For further details on \rkhs Fourier features, see~\Cref{appsec:vff}.


\subsection{Inter-domain Deep Gaussian Processes with RKHS Fourier Features}
\label{sec:dgp_rkhsfourierfeatures}

To construct inter-domain \dgps that leverage global structure in data, we use approximate posterior predictive distributions based on \rkhs Fourier Features at every layer.
For layers $\ell = 1, ..., L$ with input dimensions $D^{(\ell-1)}$, let $\omega_m = \frac{2 \pi m}{b-a}$ for $ m = 1,..., M$, and let 
\begin{align*}
    \bm{\Omega}^{(\ell - 1)} \defines [\bm{\omega}^{(\ell - 1)}_1, ..., \bm{\omega}^{(\ell - 1)}_M ]^\top   
\end{align*}
be the matrix of $M \times D^{(\ell - 1)}$ inducing frequencies producing a set of $D^{(\ell - 1)}$ truncated Fourier bases $\bm{\phi}^{(\ell)}(\fbf^{(\ell-1)})$, as defined in~\Cref{eq:fourierbasis}.
Each $ \bm{\phi}^{(\ell)}(\fbf^{(\ell-1)}) $ then maps $f^{(\ell)}$ into Fourier space by applying the \rkhs inner product $\langle \cdot , \cdot \rangle_\mathcal{H}$ given by 
\begin{align*}
    u^{(\ell)}_m  \defines \mathcal{P}_{\phi^{(\ell)}_m} (f^{(\ell)}) = \langle \bm{\phi}^{(\ell)}_m , f^{(\ell)} \rangle_\mathcal{H},
\end{align*}
for Fourier basis entries $\bm{\phi}^{(\ell)}_m(\fbf^{(\ell-1)})$ with $m = 1,..., M^\prime$ and $M^\prime = 2M + 1$ (as in the shallow \gp case), thus creating the $M^\prime \times D^{(\ell)}$-dimensional matrix
\begin{align*}
    \ubf^{(\ell)} = [ \mathcal{P}_{\phi^{(\ell)}_1}(f), ..., \mathcal{P}_{\phi^{(\ell)}_{M^\prime}}(f) ]^\top.
\end{align*}
We thus obtain inter-domain operators  $\mathbf{k}^\phi_{\mathbf{u}^{\ell}}(\fbf^{(\ell-1)})$ and $\mathbf{K}^\phi_{\mathbf{u^{\ell}u^{\ell}}}$ for \dgp layers $\ell=1, ..., L$.

Using the variational distribution in~\Cref{eq:dgp_posterior_Predictive_layer}, we then get a final-layer posterior predictive distribution
\begin{align*}
    q(\fbf_n^{(L)}) = \int \prod_{\ell =1}^{L-1} q(\fbf_n^{(\ell)} | \bm{\mu}^{(\ell)}, \bm{\Sigma}^{(\ell)}; \fbf_n^{(\ell - 1)}, \bm{\Omega}^{(\ell - 1)}) \dfbf_n^{(\ell)},
\end{align*}
where $\fbf_n^{(\ell)}$ is the $n$th row of $\fbf^{(\ell)}$.
This quantity is easy to compute using the reparameterization trick, which allows for sampling from the $n$th instance of the variational posteriors across layers by defining
\begin{align}
\label{eq:reparameterizationtrick}
    \hat{\fbf}_n^{(\ell)} = \widetilde{\mathbf{m}}\left(\hat{\fbf}_n^{(\ell - 1)}\right) + \bm{\epsilon}_n^{(\ell)} \odot \sqrt{\widetilde{\mathbf{S}} \left(\hat{\fbf}_n^{(\ell - 1)}, \hat{\fbf}_n^{(\ell - 1)}\right) }
\end{align}
and sampling from $\bm{\epsilon}_n^{(\ell)} \sim \mathcal{N}(\mathbf{0},\mathbf{I}_{D^{(\ell)}})$ \citep{Kingma2014,Salimbeni2017}.

\paragraph{Prediction}
To make predictions, we sample from the approximate posterior predictive distribution of the final layer the same way as in \dsvi.
For a test input $\xbf^\ast$, we draw $S$ samples from the posterior predictive distribution
\begin{align}
\label{eq:posterior_predictive}
    q ( \mathbf { f } _ { * } ^ { (L) } ) \approx \frac { 1 } { S } \sum _ { s = 1 } ^ { S } q(\fbf_{*}^{(L)} | \bm{\mu}^{(L)}, \bm{\Sigma}^{(L)} ; {\mathbf { f } _ { * } ^ { (s) } }^ { (L-1) } , \bm { \Omega } ^ { (L - 1) } )
\end{align}
where $q ( \mathbf { f } _ { * } ^ { (L) } )$ is the \dgps marginal distribution at $\xbf_\ast$ and ${\mathbf { f } _ { * } ^{(s)} } ^ { (L - 1) } $ are draws from the penultimate layer (and thus indirectly from all previous layers) obtained via reparameterization of each layer as shown in~\Cref{eq:reparameterizationtrick}.

\paragraph{Evidence Lower Bound}
The evidence lower bound (\elbo) is the same as in \dsvi, apart from the fact that it is computed from the inter-domain posterior predictive distributions at each \dgp layer.
It is given by
\begin{align}
\begin{split}
\label{eq:elbo}
    \calL &= \sum_{n=1}^N \E_{q(\fbf_n^{(L)})} \big[ \log p(\ybf_n | \fbf_n^{(L)}) \big] 
    \\
    &\qquad - \sum_{\ell = 1}^{L} \text{KL}(q(\ubf^{(\ell)}) \, \vert \vert \,  p(\ubf^{(\ell)})),
\end{split}
\end{align}
which can be optimized variationally using gradient-based stochastic optimization.
We include a derivation of this bound in~\Cref{appsec:elbo}.
To estimate the expected log-likelihood, we generate predictions at the input locations by drawing Monte Carlo samples from  $q(\fbf_n^{(L)})$ as shown in~\Cref{eq:posterior_predictive}.

\subsection{Further Model Details}
\label{sec:model_details}

In our implementation, we let
\begin{align}
\begin{split}
    &\bm{\omega}_{m,i}^{(\ell - 1)} = \bm{\omega}_{m,j}^{(\ell - 1)} \, \forall i, j \leq D^{(\ell - 1)} \\
    \bm{\omega}_{m}^{(\ell)} = &\bm{\omega}_{m}^{(\ell^\prime)} \,\ \forall \ell, \ell^\prime \in \{1,...,L \} \, \forall m \in \{1, ..., M^\prime \},
\end{split}
\end{align}
which means that we use the same inducing frequencies at every \dgp layer, but this assumption can easily be relaxed.
Moreover, we use additive kernels to apply \rkhs Fourier features to multidimensional inputs.
For for each layer, we define
\begin{align}
\begin{split}
    &f^{(\ell)}(\fbf^{(\ell-1)}) = \sum_{d=1}^{D^{\ell}} f^{(\ell)}_d (\fbf^{(\ell-1)}_d)
    \\
    f^{(\ell)}_d &\sim \mathcal{GP}\left(0, k^{(\ell)}_d\bigg(\fbf^{(\ell-1)}_d, {\fbf^{(\ell-1)}_d}'\bigg)\right),
\end{split}
\end{align}
where $\fbf^{(\ell-1)}_d$ is the $d$th element of the multi-dimensional single input $\fbf^{(\ell-1)}_n$, and $k^{(\ell)}_p(\cdot, \cdot)$ is a kernel defined on a scalar input space \citep{Hensman2016}.
This way, we obtain the \dgp layer
\begin{align}
    f^{(\ell)} \sim \mathcal{GP}\left(0, \sum_{d=1}^{D^{\ell}}k^{(\ell)}_d\bigg(\fbf^{(\ell-1)}_d, {\fbf^{(\ell-1)}_d}'\bigg)\right)
\end{align}
for which we are then able to construct a matrix of features with elements $u^{(\ell)}_{ m , d } = \mathcal { P } _ { \phi _ { m } }( f^{(\ell)}_{ d })$, resulting in a total of $2MD^{(\ell)}+1$ inducing variables, independent across dimensions, i.e., $\operatorname { cov } ( u^{(\ell)} _ { m , d } , u^{(\ell)} _ { m , d ^ { \prime } } ) = 0$.
With the corresponding variational parameters estimated via gradient-based optimization, the cost per iteration when computing the posterior mean for an additive kernel is $\mathcal{O}(NM^2D)$.
Using an additive kernel at each \dgp layer then results in a time complexity of \mbox{$\calO(NM^2(D^{(1)}+D^{(2)}+...+D^{(L)}))$} per iteration, which is identical to that of \dsvi.
In practice, however, we find that inter-domain \dgps require fewer inducing points and fewer gradient steps to achieve a given level of predictive accuracy compared to \dgps with \dsvi, making them more computationally efficient.

Unlike \dgps that use conventional inducing points-based approximate inference, inter-domain \dgps have an additional hyperparameter; the frequency interval $[a,b]$.
To avoid undesirable edge effects in the \dgp posterior predictive distributions, we normalize all input data dimensions to lie in the interval $[0,1]$ and define the \rkhs over the interval $[a,b] = [-2, 3]$.
We repeat this normalization at each \dgp layer before feeding the samples into the next \gp.
To avoid pathologies in \dgp models investigated in prior work \citep{Duvenaud2014}, we follow \citet{Salimbeni2017} and use a linear mean function $m^{(\ell)}(\fbf^{(\ell-1)}) = \fbf^{(\ell-1)} \mathbf{w}^{(\ell)}$, where $\mathbf{w}^{(\ell)}$ is a vector of weights, for all but the final-layer \gp, for which we use a zero mean function.
We used a Mat\'ern-$\frac{3}{2}$ kernel for all experiments.

\begin{figure*}[ht!]
\centering
\begin{subfigure}{0.31\linewidth}
  \centering
  \includegraphics[width=\columnwidth]{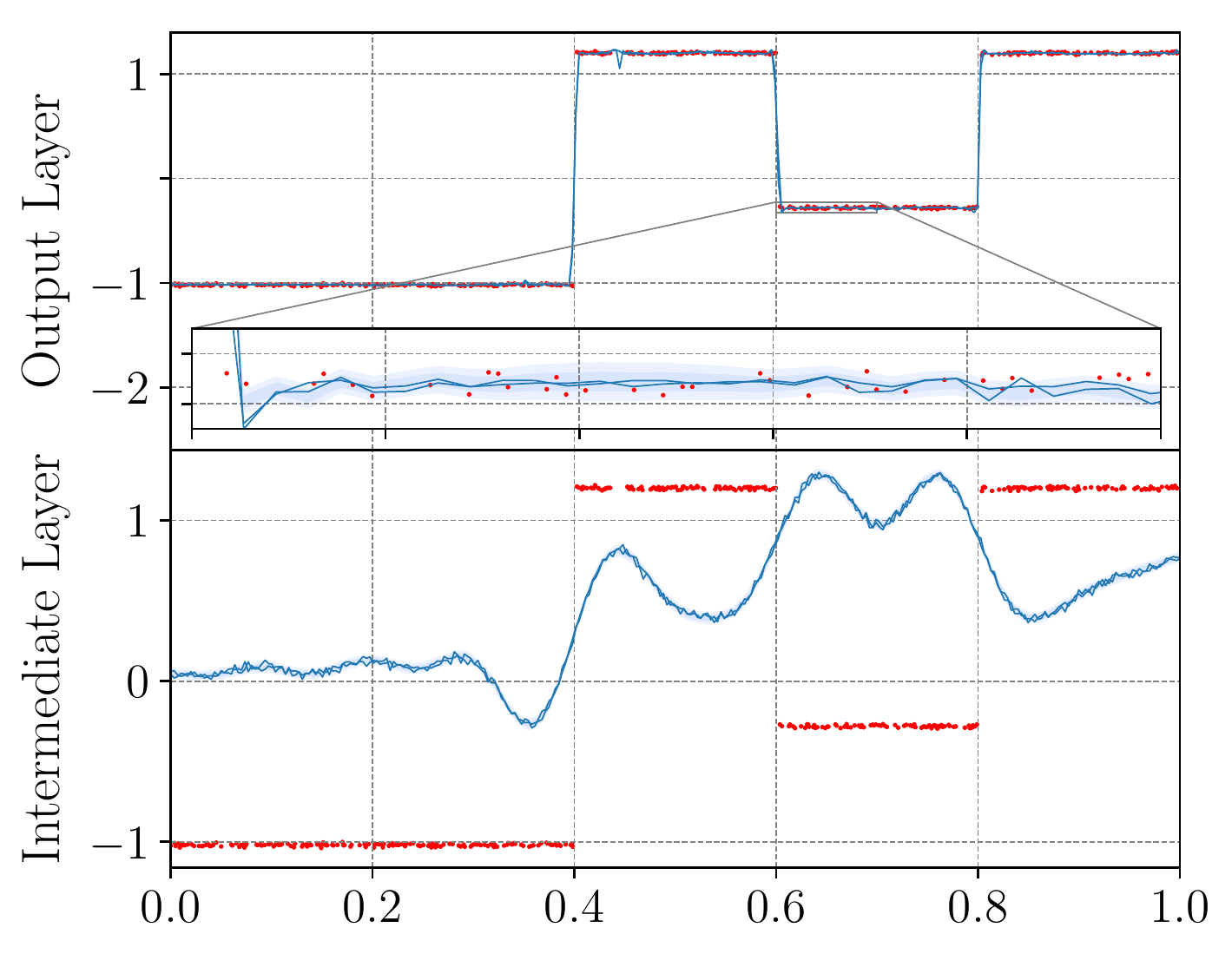}
  \caption{
    Inter-domain \dgp with \dsvi (two layers). Top: \dgp posterior predictive distribution. Bottom: Predictive distribution at intermediate layer.
  }
  \label{fig:step-iddgp}
\end{subfigure}
~\hspace{0.01cm}
\begin{subfigure}{0.31\linewidth}
  \centering
  \includegraphics[width=\columnwidth]{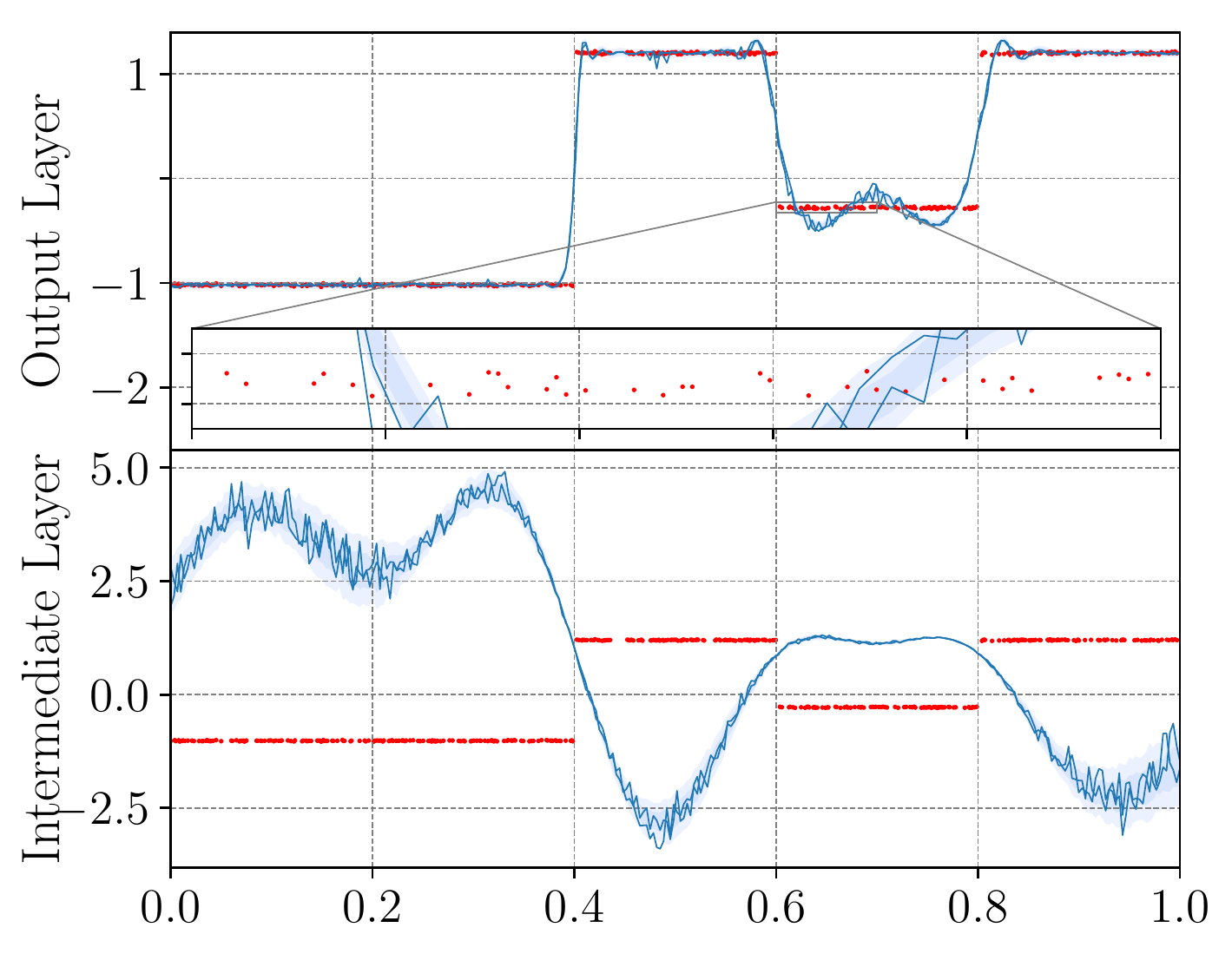}
  \caption{
    Conventional \dgp with \dsvi (two layers). Top: \dgp posterior predictive distribution. Bottom: Predictive distribution at intermediate layer.
  }
  \label{fig:step-dsvdgp}
\end{subfigure}
~\hspace{0.01cm}
\begin{subfigure}{0.301\linewidth}
  \centering
  \includegraphics[width=\columnwidth]{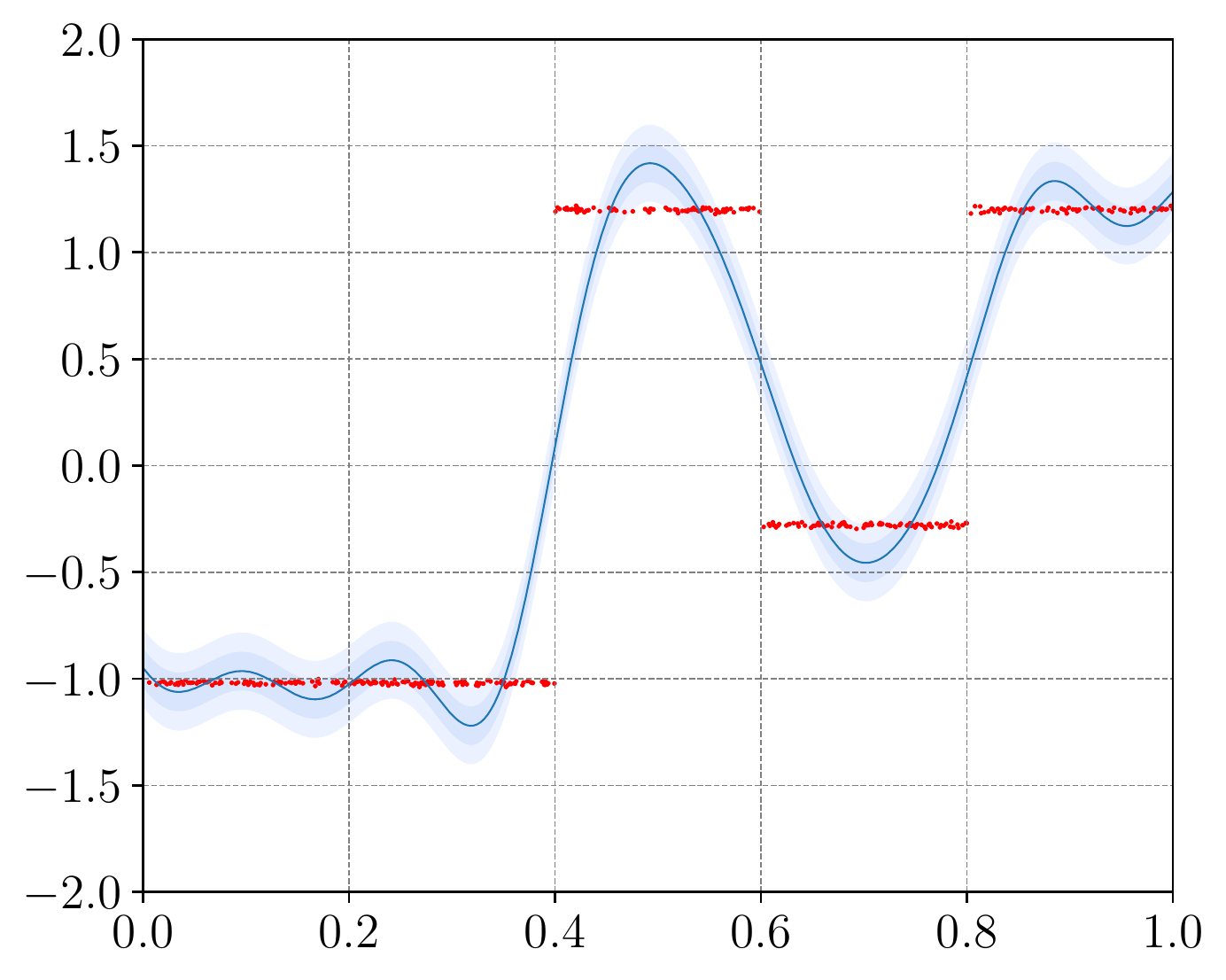}
  \caption{
      \gp with \rkhs Fourier features (single layer). Posterior predictive distribution. \\ \\
  }
  \label{fig:step-vff}
\end{subfigure}
\caption{
    Comparison of posterior predictive distributions of different \gp models on synthetic non-stationary data.
    The models are trained using 20 inducing frequencies and 20 inducing points, respectively.
    In each plot, training points are shown in red.
    Each shade of blue represents one standard deviation in the posterior predictive distribution.
    For enlarged plots, see~\Cref{appsec:experiments}.
}
\label{fig:multi_step_function}
\end{figure*}


\section{Related Work}
\label{sec:relatedwork}

Inducing points-based approximate inference has allowed \gps models to scale to large numbers of input points \citep{Snelson2005,Titsias2009,Hensman2013,Bui2014,Hensman2015a}.
Our work directly builds on \citet{Hensman2016} and \citet{Salimbeni2017} and adds to the literature on sparse spectrum approximations \citep{Lazaro-Gredilla-IDGP,Lazaro-GredillaCRF10,Gal2015,Wilson2015kernel}.
Specifically, we extend \citet{Hensman2016} to compositions \gp models by leveraging the compositional structure of the approximate posterior of \citet{Salimbeni2017}.
In contrast to \citet{Wilson2015kernel}, \citet{Lazaro-Gredilla-IDGP}, and \citet{Gal2015}, \citet{Hensman2016} (and, by extension, our approach) combines inter-domain operators with \svi \citep{Hensman2013} and is amenable to stochastic optimization on minibatches, which makes it possible to apply it to large datasets without facing memory constraints.
Similar to our approach, random feature expansions for \dgps \citep{Cutajar2017} use projections of each \dgp layer's predictive distribution onto the spectral domain to perform approximate inference, but unlike our approach, it is not based on inducing points.


\section{Empirical Evaluation}
\label{sec:experiments}

To demonstrate that inter-domain \dgps improve upon inter-domain shallow \gps in their ability to model complex, non-stationary data and to show that inter-domain \dgps improve upon local inducing points-based approximate inference methods for \dgps, we will present results from several experiments that showcase the types of prediction problems for which inter-domain \dgps are particularly well-suited.
We are particularly interested in modeling complex data-generating processes which exhibit global structure as well as non-stationarity, since the former is challenging for \dgps that use local approximations, such as \dsvi for \dgps, and the latter is challenging for shallow \gps with stationary covariance functions.

To illustrate the advantage of inter-domain deep \gps over inter-domain shallow \gps in modeling non-stationary data, we present a suite of qualitative and quantitative empirical evaluations on datasets that exhibit global structure and non-stationarity.

First, we present a simple, synthetic data experiment designed to demonstrate that our method is well-suited for modeling data from generating processes that exhibit both non-stationarity and global structure.
Next, we illustrate that inter-domain deep \gps provide a significant gain in computational efficiency when modeling data that exhibits global structure.
In particular, we compare the number of inducing frequencies and inducing points needed to attain a certain predictive accuracy when using inter-domain \dgps and local inducing points-based \dgps on a challenging real-world audio sub-band reconstruction task.
Lastly, we demonstrate that our method outperforms existing state-of-the-art shallow \gps with local approximate inference, shallow \gps with global approximate inference, and deep \gps with local approximate inference on a series of challenging real-world benchmark prediction tasks.
For additional experiments and more experimental details, see~\Cref{appsec:experiments}.

\begin{figure*}[ht!]
\centering
    \includegraphics[width=\textwidth]{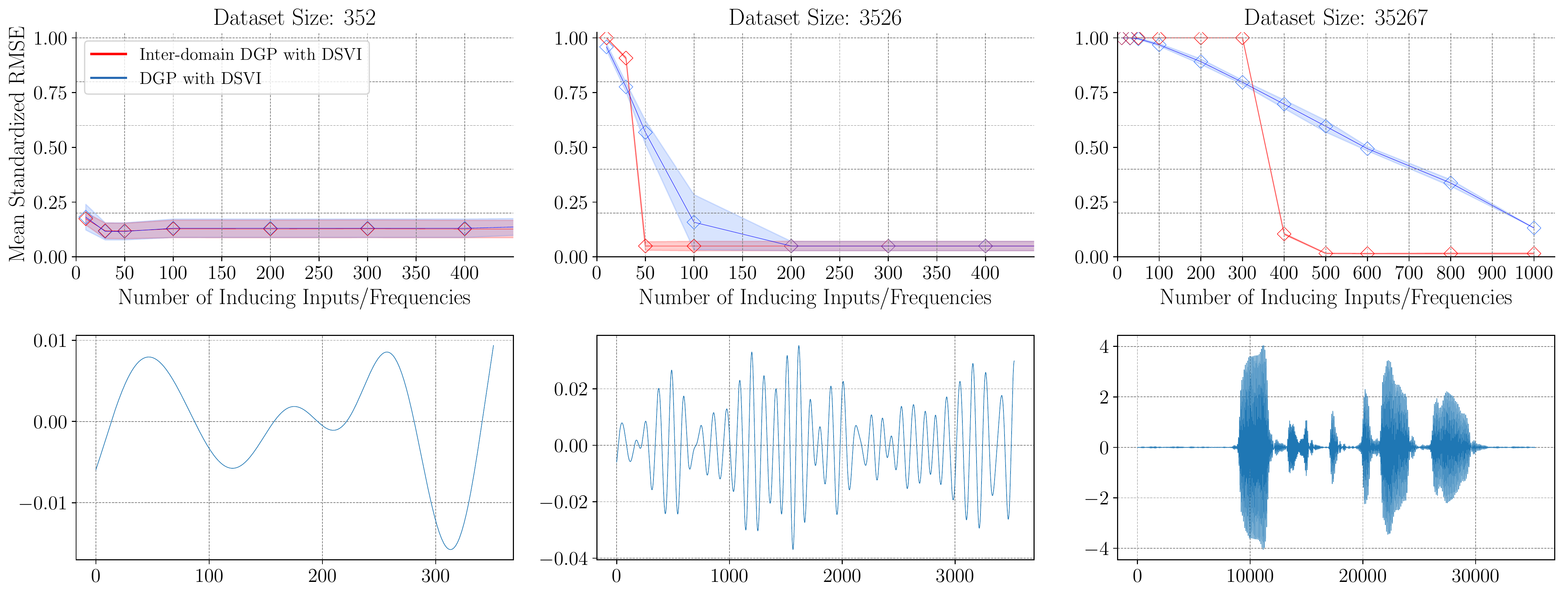}
\caption{
    Comparison of average standardized root mean squared errors for varying numbers of inducing inputs on three datasets of increasing global structure and complexity.
    On complex datasets (center and right panel), Inter-domain deep \gps with \dsvi require fewer inducing inputs than conventional \dgps with \dsvi.
    Standardized root mean squared errors were evaluated on a test set of 40\% of datapoints in each subset over 10 random seeds each.
}
\label{fig:audio_comp}
\end{figure*}


\subsection{Highly Non-Stationary Data with Global Structure}

The multi-step function in~\Cref{fig:multi_step_function} is designed to exhibit \textit{both} global structure \textit{as well as} non-stationarity, providing an optimal test case to assess the performance of inter-domain \dgps vis-\`a-vis related methods on a simple and easily interpretable prediction task.

The plot shows the posterior predictive distributions of inter-domain \dgps, \dgps with \dsvi, and inter-domain shallow \gps with \rkhs Fourier features.
As can be seen in the plots, inter-domain \dgps are the only method that is able to model the step locations well and to infer the global structure---that is, that the function is constant within certain intervals---with high accuracy and good predictive uncertainty--despite having a stationary covariance function (see~\Cref{fig:step-iddgp}).
Inter-domain shallow \gps, in contrast, are unable to capture either the step transitions nor the global structure, reflecting their limited expressiveness (see~\Cref{fig:step-vff}).
While \dgps benefit from increased expressivity, they, too, fail to fully capture the global structure and the non-stationarity (see~\Cref{fig:step-dsvdgp}).
This is due to the inherently local nature of local inducing points-based inference, which requires large numbers of inducing inputs to accurately approximate complex posterior distributions.
The experiment illustrates that inter-domain \dgps are in fact able to overcome key limitations of both shallow \gp inter-domain approaches and outperform state-of-the-art local \dgp inference methods.

\Cref{fig:multi_step_function} highlights another key difference between conventional and inter-domain \dgps.
In particular, the bottom plots in~\Cref{fig:step-iddgp} and~\Cref{fig:step-dsvdgp}, show the output of the \dgp intermediate layers and are markedly different from one another.
While it appears that the conventional \dgp seeks to model the target function directly in intermediate-layer space, the inter-domain \dgp appears to cluster datapoints from the original input space in a way so that the changes in the step function in output space become associated with smooth transitions in intermediate-layer space.

\begin{figure*}[ht!]
\centering
    \includegraphics[width=\textwidth]{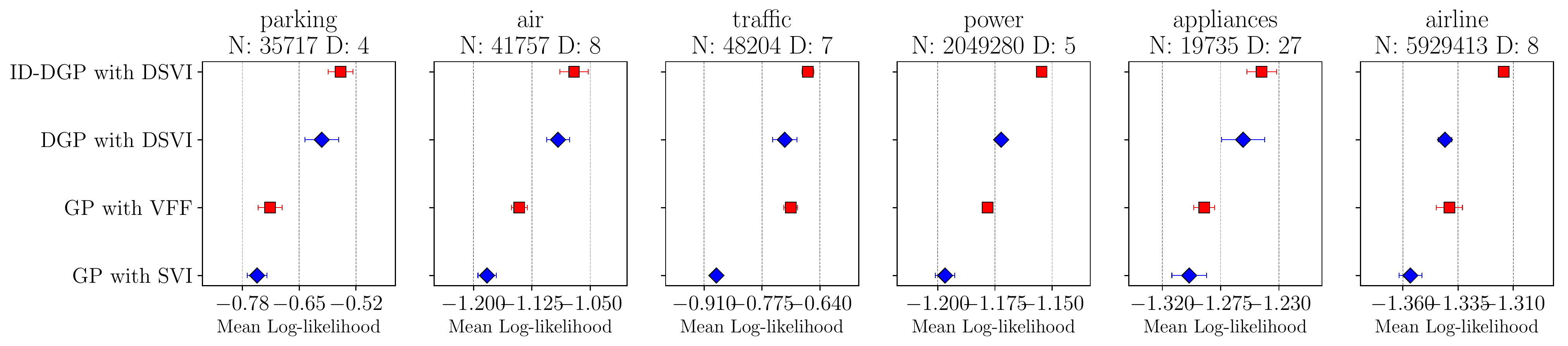}
\caption{
    Average test log-likelihood (higher is better) and standard errors (over 10 random seeds) on a set of real-world datasets with global structure.
    All models were trained with 50 inducing points.
    The inter-domain \dgp with \dsvi has two layers and the conventional \dgp with \dsvi has four layers.
    The performance of the inter-domain \dgp did not increase as additional layers were added.
}
\label{fig:benchmarks-llk}
\end{figure*}


\subsection{Leveraging Global Structure to Effectively and Efficiently Model Complex Data}

Next, we quantitatively assess the predictive accuracy and computational efficiency of inter-domain \dgps.
To do so, we use a smoothed sub-band of a speech signal taken from the TIMIT database and previously used in \citet{Bui2014}.
The dataset exhibits complex global structure which is difficult to model using local approximation methods.
To assess how well inter-domain \dgps are able to capture global structure in the data, we compare it to a doubly stochastic variational inference for \dgps, a state-of-the-art approximate inference method for \dgps based on local inducing points.
To assess how well different approximate inference methods are able to capture the complex global structure, we look at three subset of the data:
the first 352, 3,526, and 35,267 datapoints, respectively.

The smallest subset of only 352 datapoints does not exhibit much global structure and is small enough to be modeled with few (local) approximations, which is reflected by the results in the left panel of~\Cref{fig:audio_comp}, where inter-domain \dgps and conventional \dgps perform equally well, and increasing the  number of inducing frequencies/points does not lead to an improvement in performance.
For the larger dataset containing 3,526 datapoints, however, the global structure--indicated by a high degree of autocorrelation in the data--becomes readily apparent.
As can be seen in the top row, inter-domain \dgps require relatively fewer inducing frequencies compared to conventional \dgps to achieve a test error close to zero.
The difference in the number of inducing points required to model the data is most significant for the largest subset, shown in the right panel of~\Cref{fig:audio_comp}.
As can be seen in the plot, the covariance of the process varies significantly.
This subset of the audio sub-band dataset is highly non-stationary and exhibit global structure in the form of high autocorrelation.
As a result, inter-domain deep \gps are able to attain a test error close to zero with fewer than half the number of inducing points needed to achieve the same level of accuracy with conventional \dgps.

Since the time complexity of \dsvi scales quadratically in the number of inducing points and inter-domain and conventional \dgps have the same time complexity (that is, a single gradient step takes approximately equally long for the same number of inducing frequencies/points), inter-domain \dgps are more computationally efficient in practice when modeling data exhibiting global structure.
Additionally, in~\Cref{fig:audio-pred}, we also show that for 20 inducing frequencies/points, inter-domain \dgps have better-calibrated posterior predictive uncertainty estimates than conventional \dgps.


\subsection{Global Structure in Real-World Data}

To quantitatively assess the predictive performance of inter-domain \dgps, we evaluate them on a range of real-world dataset, which exhibit global structure---usually in the form of a temporal component that induces a high autocorrelation.
The experiments include medium-sized datasets (`parking', `air', `traffic'), two very large datasets with over two and five million datapoints each (`power' and `airline'), and a high-dimensional dataset with 27 input dimensions (`appliances').
As can be seen in~\Cref{fig:benchmarks-llk}, inter-domain \dgps consistently outperform conventional \dgps (\dgps with \dsvi) as well as inter-domain shallow \gps (\gps with \vff) and significantly outperform conventional shallow \gps (\svi), suggesting that combining the increased expressivity of \dgp models with the ability of inter-domain approaches to capture global structure leads to the best predictive performance.
See~\Cref{appsec:experiments} for a plot of the test standardized RMSEs for the experiments in~\Cref{fig:benchmarks-llk} and for additional results on datasets that do not exhibit global structure (and on which our method performs on par with existing methods).

To assess the predictive performance of inter-domain \dgps on extremely complex, non-stationary data, we test our method on the U.S. flight delay prediction problem, a large-scale regression problem that has reached a status of a standard test in \gp regression due to its massive size of $5,929,413$ observations and its non-stationary nature, which makes it challenging for \gps with stationary covariance functions \citep{Hensman2016}.
The data set consists of flight arrival and departure times for every commercial flight in the United States for the year 2008.
We predict the delay of the aircraft at landing (in minutes) from eight covariates: the age of the aircraft (number of years since deployment), route distance, airtime, departure time, arrival time, day of the week, day of the month, and month.
The non-stationarity in the data is likely due to the recurring daily, weekly, and monthly fluctuations in occupancy.
In our evaluation, we find that the predictive performance of inter-domain \dgps is superior to closely-related state-of-the-art shallow and deep \gps as shown in~\Cref{tab:res_airline} and~\Cref{fig:benchmarks-llk}.

\begin{table}[h!]
    \caption{
        Average standardized root mean squared errors and standard errors (over 10 random seeds) on the U.S. flight delay prediction task.
    }
    \centering
\begin{tabular}{@{\extracolsep{-8pt}}lcc}
\hline \\[-1.8ex]
$N$ & \multicolumn{1}{c}{1,000,000} & \multicolumn{1}{c}{5,929,413}  \\
Method & \multicolumn{1}{c}{RMSE $\pm$ SE} & \multicolumn{1}{c}{RMSE $\pm$ SE}  \\
\hline  \\[-2ex]
\textsc{id-dgp} with \dsvi  & $ \mathbf{ 0.906 \pm 0.006 } $ & $ \mathbf{0.903 \pm 0.002}$ \\
\dgp with \dsvi $~~~~~~~~~~$  & $\, 0.932 \pm 0.004 $ & $ 0.930 \pm 0.003 $ \\
\gp with \vff $~~~~~~~~~~$ & $\, 0.925 \pm 0.007 $ & $ 0.923 \pm 0.006$  \\
\gp with \svi        & $\, 0.946 \pm 0.008  $ & $0.941 \pm 0.005 $  \\
\hline \\ [-1.8ex] 
\end{tabular} \label{tab:res_airline}
\end{table}


\section{Conclusion}

We proposed \textit{Inter-domain Deep Gaussian Processes} as a deep extension of inter-domain \gps that combines the advantages of inter-domain and deep \gps and allows us to model data exhibiting non-stationarity \textit{and} global structure with high predictive accuracy and low computational overhead.
We showed how to leverage the compositional nature of the approximate posterior in \dsvi to perform simple and scalable approximate inference and established that inter-domain \dgps can be more computationally efficient than conventional \dgps.
Finally, we demonstrated that our method significantly and consistency outperforms inter-domain shallow \gps and conventional \dgps on data exhibiting non-stationarity and global structure.


\clearpage

\section*{Acknowledgements}

Tim G. J. Rudner is funded by the Rhodes Trust and the Engineering and Physical Sciences Research Council (EPSRC).
We would like to thank Stephen Roberts, James Hensman, Andreas Damianou, Zhenwen Dai, and Neil Lawrence for helpful discussions.


\bibliography{references}
\bibliographystyle{include/icml2020}


\clearpage

\begin{appendices}
\onecolumn

\crefalias{section}{appsec}
\crefalias{subsection}{appsec}
\crefalias{subsubsection}{appsec}

\setcounter{equation}{0}
\renewcommand{\theequation}{\thesection.\arabic{equation}}


\section*{\LARGE Supplementary Material}
\label{appsec:appendix}

\medskip


\section{Project Website}

For source code and additional results, see \url{https://bit.ly/inter-domain-dgps}.


\section{Further Experiments \& Experimental Details}
\label{appsec:experiments}

In addition to the experiments presented in~\Cref{sec:experiments}, we performed several qualitative and quantitative evaluations to better understand the properties and training behavior of \textit{Inter-domain} \dgps.

\begin{figure}[h!]
\centering
    \includegraphics[width=\textwidth]{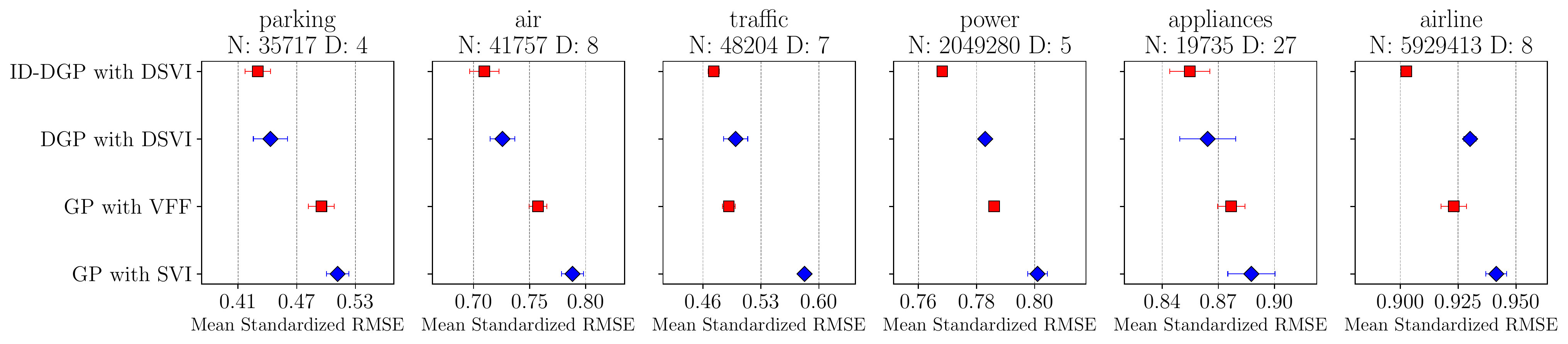}
\caption{Average standardized root mean squared error (lower is better) and standard errors (over 10 random random seeds) on a set of real-world datasets exhibiting global structure. All models were trained with 50 inducing points.}
\label{fig:benchmarks-rmse}
\end{figure}

In this section, we include plots that provide insights into the effect of adding additional layers to an Inter-domain \dgp (see~\Cref{fig:step-iddgp} and~\Cref{fig:step-dsvdgp}) and plot draws from the inter-domain and conventional \dgp priors.
We also include plots of the posterior predictive distributions of inter-domain deep \gps, conventional \dgps, and inter-domain shallow \gps (see~\Cref{fig:audio-pred}) as well as standardized RMSEs for the real-world experiments presented in~\Cref{fig:benchmarks-llk} in the main paper (see~\Cref{fig:benchmarks-rmse}).
Furthermore, we include average test root mean squared errors and log-likelihoods for a selection of datasets that do \textit{not} exhibit global structure.

\subsection{Training Details}

\textbf{Model}$\,\,\,$ For \textsc{dsvi-dgp}{\normalfont s}, we set the number of hidden units per layer equal to the number of input dimensions. We used both the RBF and the Mat\'ern-$\frac{3}{2}$ kernels with automatic relevance determination (\textsc{ard}) for all experiments, but models with RBF kernel performed better.

\textbf{Training}$\,\,\,$ For the benchmark deep \gp models, we used learning rates suggested by the authors as well as $10^{-2}$ and $10^{-3}$ for all experiments. For inter-domain and conventional \dgps with \dsvi, we used $\textit{Adam}$ optimizer with learning rates of $10^{-2}$ and $10^{-3}$ for all regression tasks. For benchmark shallow \gp models, we used the \textsc{BFGS} algorithm. For all other models, we used the implementation default optimizer.

\textbf{Parameter initializations}$\,\,\,$ For both inter-domain and conventional \dgps with \dsvi, we initialized the inducing function value means to zero and variances to the identity and $10^{-5}$ for outer and inner layers, respectively. For all inducing points-based methods, we initialized the inducing inputs using the K-means algorithm on the training inputs.


\subsection{Experiments}

\begin{figure}[h!]
    \centering
    \includegraphics[width=\textwidth]{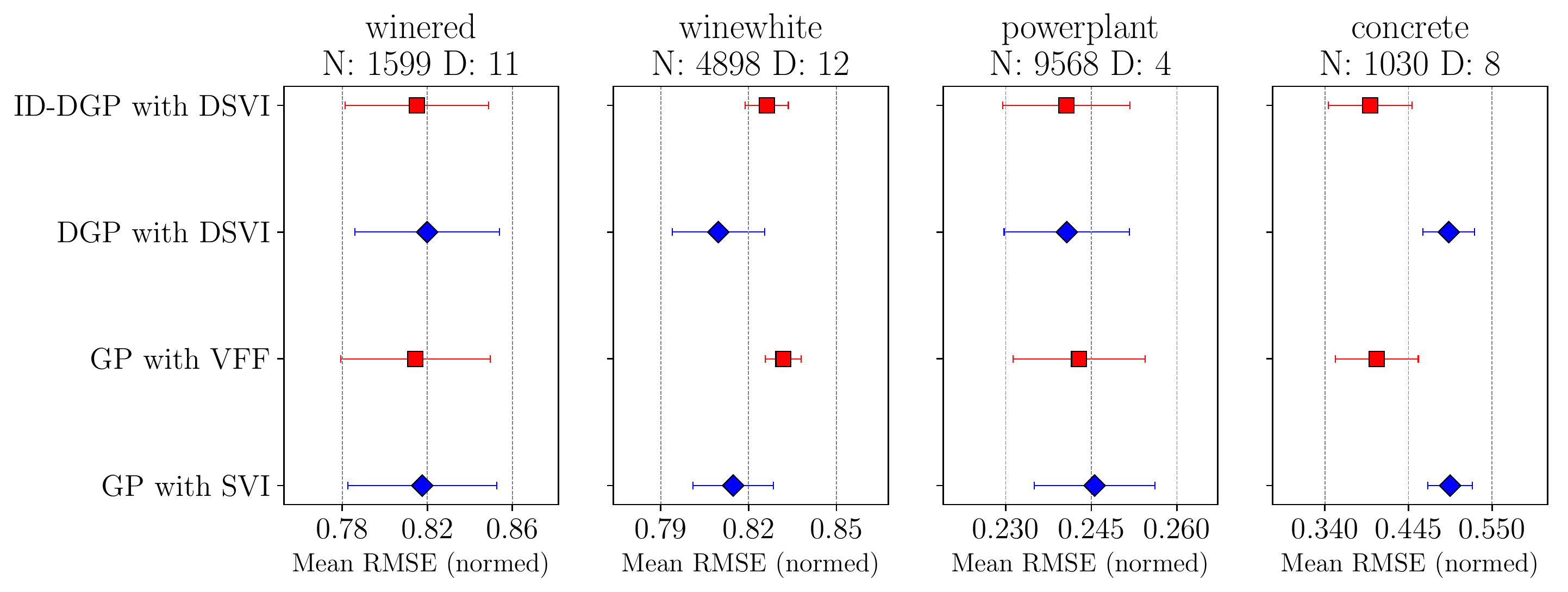}
    \caption{
        Average standardized root mean squared errors (lower is better) and standard errors (over 10 random seeds) on a set of small- and medium-scale regression problems. Each model was trained with 20 inducing points/inducing frequencies.
    }
    \label{fig:uci-rmse}
\end{figure}

\begin{figure}[h!]
    \centering
    \includegraphics[width=\textwidth]{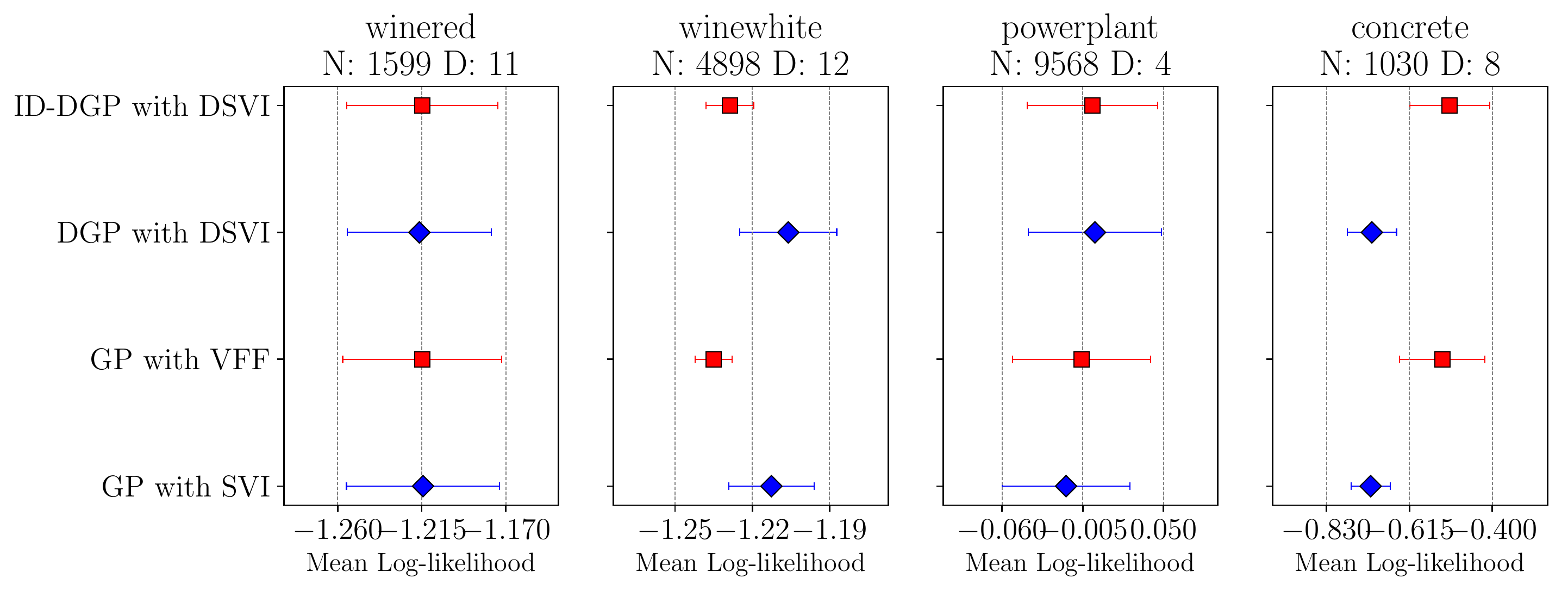}
    \caption{
        Average test log-likelihoods (higher is better) and standard errors (over 10 random seeds) on a set of small- and medium-scale regression problems. Each model was trained with 20 inducing points/inducing frequencies.
    }
    \label{fig:uci-llk}
\end{figure}

\clearpage

\begin{figure}[h!]
    \centering
    \includegraphics[width=\textwidth]{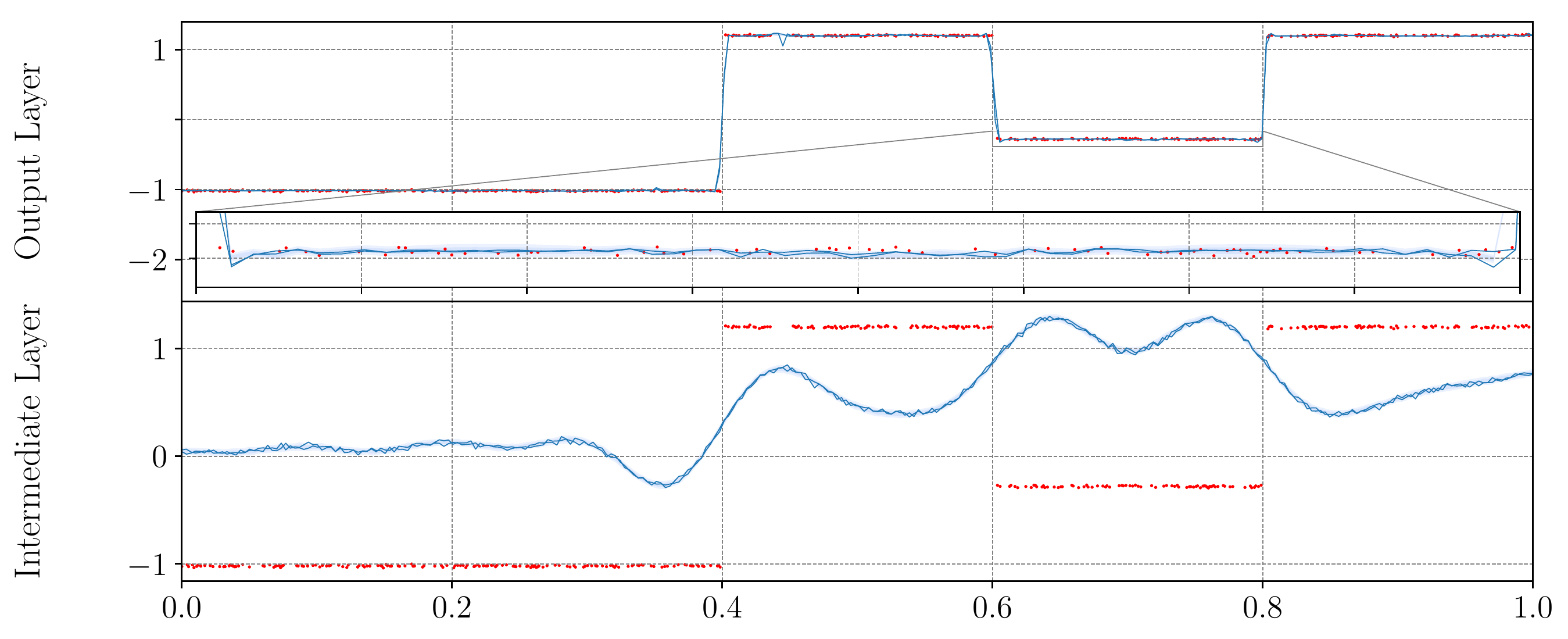}
    \caption{
        Inter-domain \dgp with \dsvi (two layers). Top: \dgp posterior predictive distribution. Bottom: Marginal distribution at intermediate layer. The model is trained using 20 inducing frequencies. Training points are shown in red. Each shade of blue represents one standard deviation in the posterior predictive distribution.
    }
    \label{fig:step-iddgp-wide}
\end{figure}

\begin{figure}[h!]
    \centering
    \includegraphics[width=\textwidth]{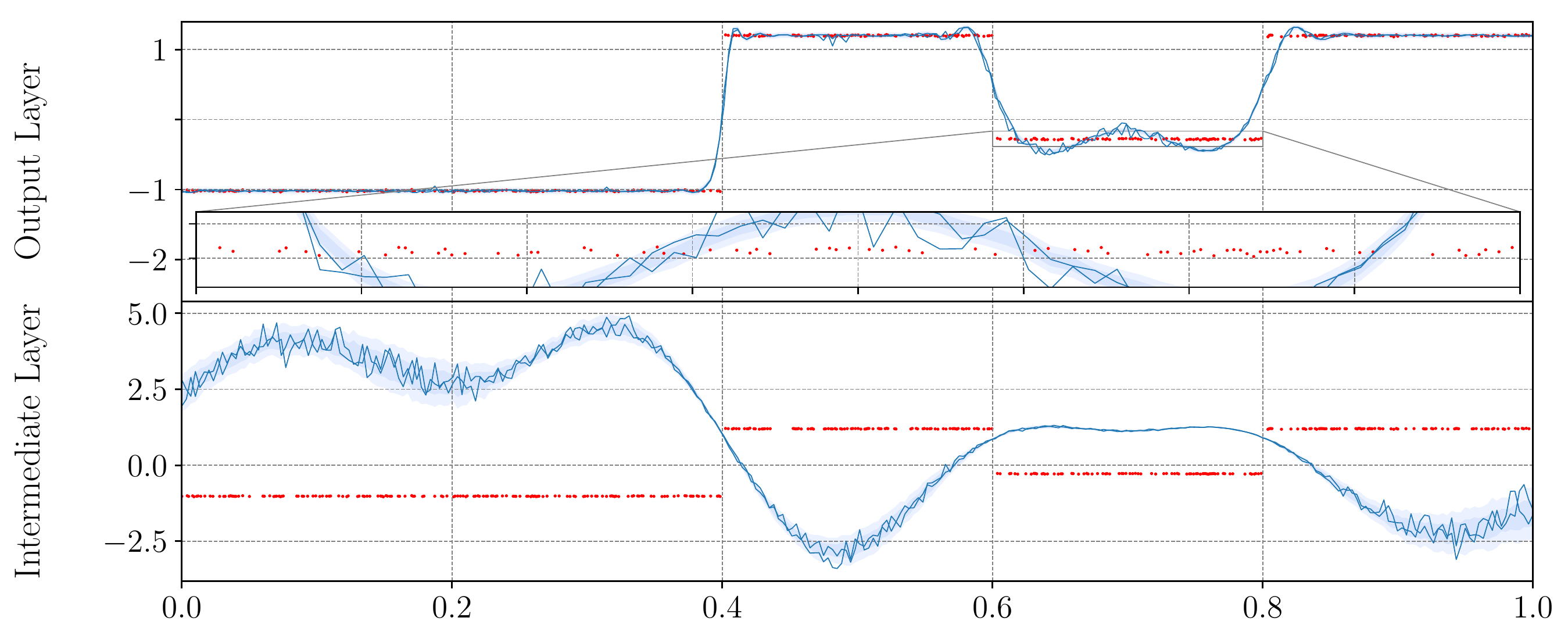}
    \caption{
        Conventional \dgp with \dsvi (two layers). Top: \dgp posterior predictive distribution. Bottom: Marginal distribution at intermediate layer. The model is trained using 20 inducing frequencies. Training points are shown in red. Each shade of blue represents one standard deviation in the posterior predictive distribution.
    }
    \label{fig:step-dsvidgp-wide}
\end{figure}

\Cref{fig:step-iddgp-wide} and~\Cref{fig:step-dsvidgp-wide} are enlarged versions of the plots in the main paper.
As can be seen from the plots, Inter-domain \dgps with \dsvi outperform conventional \dgps with \dsvi in modeling global structure, here exemplified by the different plateaus in the step function.


\clearpage

\begin{figure}[h!]
    \centering
    \includegraphics[width=\textwidth]{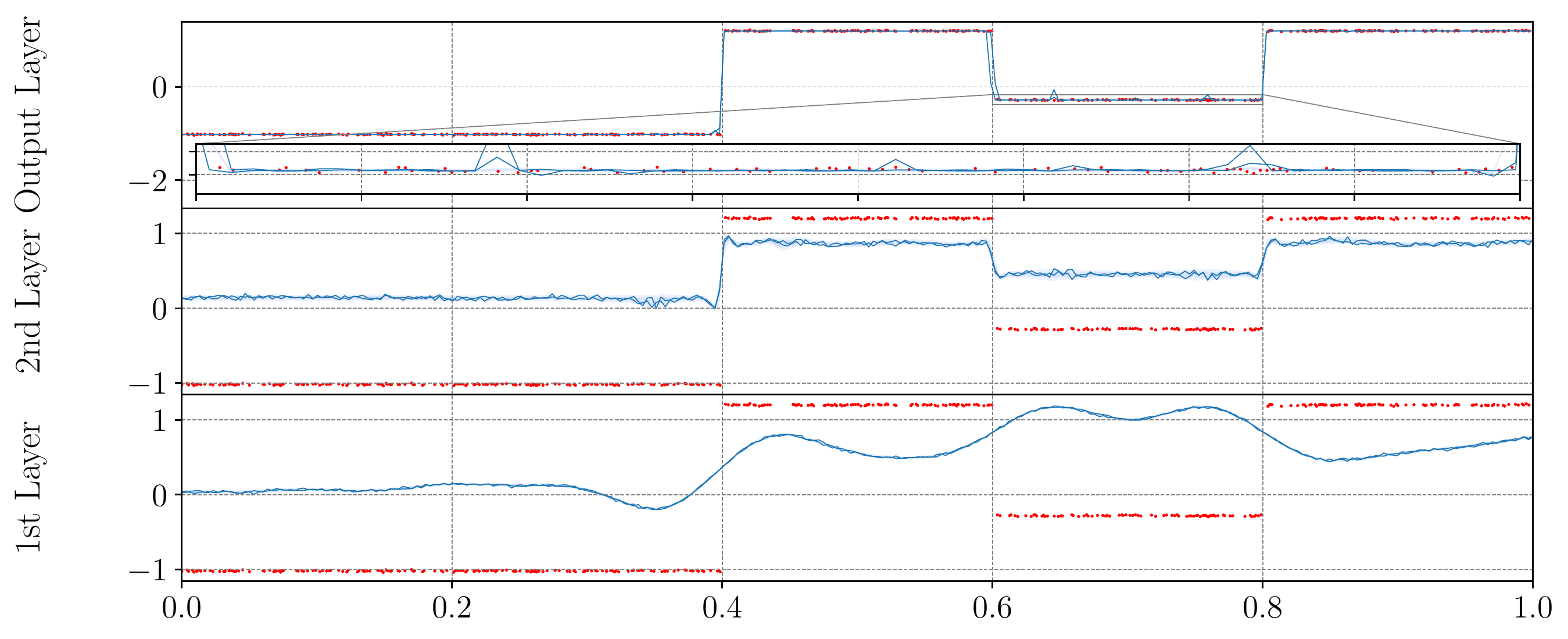}
    \caption{
        Inter-domain \dgp with \dsvi (three layers). Top: \dgp posterior predictive distribution. Bottom: Marginal distribution at intermediate layer. The model is trained using 20 inducing frequencies. Training points are shown in red. Each shade of blue represents one standard deviation in the posterior predictive distribution.
    }
    \label{fig:step-iddgp-3L-wide}
\end{figure}

\begin{figure}[h!]
    \centering
    \includegraphics[width=\textwidth]{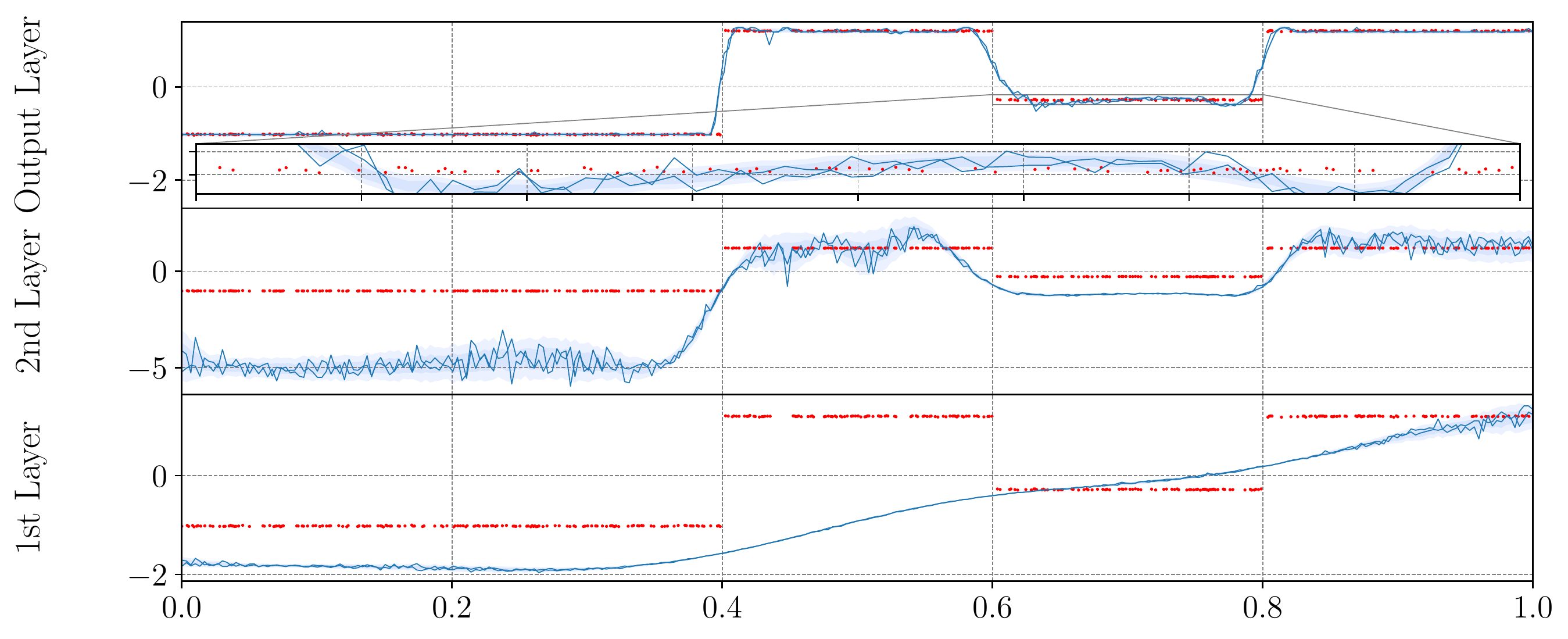}
    \caption{
        Conventional \dgp with \dsvi (three layers). Top: \dgp posterior predictive distribution. Bottom: Marginal distribution at intermediate layer. The model is trained using 20 inducing frequencies. Training points are shown in red. Each shade of blue represents one standard deviation in the posterior predictive distribution.
    }
    \label{fig:step-dsvidgp-3L-wide}
\end{figure}

\Cref{fig:step-iddgp-3L-wide} and~\Cref{fig:step-dsvidgp-3L-wide} show the outputs of individual \dgp layer for inter-domain \dgps with \dsvi and conventional \dgps with \dsvi, respectively.
As can be seen from the plots, both penultimate layers (i.e., the 2nd layers), approximately reflect the shape of the data and of the output of the final layer.
Notably, both penultimate layers appear to be (vertically) scaled versions of the output layer, which suggest that adding additional layers allow the model to `approach' the function it is trying to model slowly with each \gp composition.
Comparing the output layer predictions in~\Cref{fig:step-iddgp-wide} and~\Cref{fig:step-iddgp-3L-wide}, however, we do not observe a significant difference.
One notably difference between~\Cref{fig:step-iddgp-3L-wide} and~\Cref{fig:step-dsvidgp-3L-wide}, however, is that the first-layer output of the inter-domain \dgp is non-monotone, whereas that of the conventional \dgp roughly is.


\clearpage


\section{RKHS Fourier Features for Approximate Inference in Gaussian Processes}
\label{appsec:vff}

\rkhs Fourier features were introduced as an inter-domain representation of inducing variables in variational inference for shallow \gps by \citet{Hensman2016}.
\rkhs Fourier features use \rkhs theory to construct inter-domain alternatives to the covariance matrices $\Kuu$ and $\ku(\xbf)$ used in inducing points-based approximate inference methods. They are constructed by projecting the target function $f$ onto the truncated Fourier basis,
\begin{gather}
\begin{aligned}
    \bm{\phi}(x) =  &[1, \cos(\omega_1(x - a)), ..., \cos(\omega_M (x - a)), \sin(\omega_1 (x - a)), ..., \sin(\omega_M (x - a))]^\top,
\end{aligned}
\end{gather}
where $x$ is a single, one-dimensional input, and the $m$th frequency $\omega_m$ is defined as
\begin{align*}
    \omega_{m} = \frac{2 \pi m}{b-a}
\end{align*}
for some interval $[a,b]$. The specific functional form of the truncated Fourier basis is derived from the basis function used for \textit{Random Fourier Features} \citep{Rahimi2007}.
\citet{Berlinet2004} showed that if $\mathcal{F} = \text{span}(\bm{\phi})$ is a subspace of an \rkhs $\mathcal{H}$, the kernel of $\mathcal{F}$ is given by
\begin{align*}
    k_\mathcal{F}(x, x') = \bm{\phi}(x)^\top \mathbf{K}^{-1}_{\phi \phi} \bm{\phi}(x'),
\end{align*}
where $\mathbf{K}_{\phi \phi}[m, m'] = \langle \phi_m, \phi_{m'} \rangle_\mathcal{H}$ is the Gram matrix of $\bm{\phi}$ in $\mathcal{H}$ and $\bm{\phi}_m(x)$ is an entry of $\bm{\phi}(x)$ with $m = 1, ..., M^\prime$ and $M^\prime = 2M + 1$ for $M$ frequencies.
Furthermore, for an \rkhs $\mathcal{H}$, the coordinate of the projection of a function $h \in \mathcal{H}$ onto $\bm{\phi}_m(x)$ is given by
\begin{align*}
    \mathcal{P}_{\bm{\phi}_m}(h) = \langle h, \bm{\phi}_m \rangle_{\mathcal{H}}
\end{align*}
and defines a projection between domains. \citet{durrande-periodicities16} showed that if $\mathcal{H}$ is a Mat\'ern \rkhs of functions over $[a, b]$ with a half-integer parameter, then $\mathcal{F}$ belongs to $\mathcal{H}$. The authors also provided closed-form expressions of the inner products for the Mat\'ern-$\frac{1}{2}$, Mat\'ern-$\frac{3}{2}$, and Mat\'ern-$\frac{5}{2}$ \rkhs.
However, in order to apply the \rkhs inner product $u _ { m } = \langle f, \phi _ { m } \rangle _ { \mathcal { H } }$ between the sinusoids and the \gp sample path, which \textit{a priori} does not belong to the \rkhs, it is necessary to extend the operators $\mathcal { P } _ { \phi _ { m } } : h \mapsto \left\langle h, \phi _ { m } \right\rangle _ { \mathcal { H } }$ to square integrable functions.
\citet{Hensman2016} show that this is possible for the half-integer members of the Mat\'ern family of kernels.
With these results, we can construct the inducing variables as an inter-domain projection by letting $u_m = \mathcal{P}_{\bm{\phi}_m}(f)$, which yields
\begin{align*}
    \text{cov}(u_m, f(x)) = \bm{\phi}_m(x), \qquad \text{cov}(u_m, u_{m'}) = \langle \bm{\phi}_{m}, \bm{\phi}_{m'} \rangle_\mathcal{H},
\end{align*}
for both of which there are closed-form expressions for the half-integer members of the Mat\'ern family of kernels \citep{durrande-periodicities16}, provided in \citet{Hensman2016}. The resulting operators
\begin{align*}
    \mathbf{k}^\phi_{\mathbf{u}}(x) = \bm{\phi}_m (x), \qquad \mathbf{K}^\phi_{\mathbf{uu}} = \langle \bm{\phi}_{m}, \bm{\phi}_{m'} \rangle_\mathcal{H},
\end{align*}
represent generalized, inter-domain alternatives to the $\ku(x)$ and $\Kuu$ operators used in local inducing-points approaches. Note that, as is the case for covariance matrices in local inducing-points methods, the variational Fourier feature operators $\mathbf{k}^\phi_{\mathbf{u}}(x)$ and $\mathbf{K}^\phi_\mathbf{{u u}}$ relate model inputs to the output space, but in contrast to inducing inputs in local inducing-points approaches, the inducing frequencies do not need to lie in the same space as the model inputs.


\section{Doubly Stochastic Variational Inference for Deep Gaussian Processes}
\label{appsec:dsvi}

Inter-domain \dgps exploit the compositional structure of the approximate posterior in \textit{doubly stochastic variational inference} for \dgps to achieve simple and scalable inference in inter-domain \dgps.

In \textit{doubly stochastic variational inference}, proposed by \citet{Salimbeni2017}, the variational posterior is defined to have the following three properties:
First, conditioned on $\ubf^{(\ell)}$, the variational distribution is assumed to maintain the exact model,
\begin{align*}
    q(\fbf^{(\ell)}, \ubf^{(\ell)}) = p(\fbf^{(\ell)} | \ubf^{(\ell)}) q(\ubf^{(\ell)});
\end{align*}
second, a mean-field assumption is made so that the posterior distribution of $\{\ubf^{(\ell)}\}_{\ell = 1}^ {L}$ factorizes across layers (and dimensions), which implies that the variational distribution takes the form
\begin{align*}
    \mathcal{Q}
    &=
    q(\{ \fbf^{(\ell)} , \ubf^{(\ell)} \}_{\ell = 1}^L ) = \prod_{\ell = 1}^{L} p(\fbf^{(\ell)} | \ubf^{(\ell)}, \fbf^{(\ell - 1)}) q(\ubf^{(\ell)});
\end{align*}
and third, $q(\ubf^{(\ell)})$ is assumed to be Gaussian with mean $\bm{\mu}^{(\ell)}$ and variance $\bm{\Sigma}^{(\ell)}$ for $\ell = 1,..., L$. These properties make it possible to marginalize out the set of $\ubf^{(\ell)}$ from $\mathcal{Q}$ analytically, which yields
\begin{align}
\begin{split}
\label{eq:marginalvar}
    q(\{ \fbf^{(\ell)} \}^{L}_{\ell = 1})
    &=
    \prod_{\ell = 1}^{L} q(\fbf^{(\ell)} | \bm{\mu}^{(\ell)}, \bm{\Sigma}^{(\ell)}; \fbf^{(\ell - 1)}, \Zbf^{(\ell - 1)})
    \\
    &=
    \prod_{\ell = 1}^{L} \mathcal{N}(\fbf^{(\ell)} | \widetilde{\mathbf{m}}_{\fbf}^{(\ell)}, \widetilde{\mathbf{S}}_{\fbf}^{(\ell)} ),
\end{split}
\end{align}
where
\begin{align}
    \begin{split}
    \widetilde{\mathbf{m}}_{\fbf}^{(\ell)}
    &\defines
    \widetilde{\mathbf{m}} (\fbf^{(\ell)})
    \\
    &=
    \mathbf{m_{\fbf^{\ell}}} - \mathbf{K_{f^{\ell} u^{\ell}}} \mathbf{K}_{\mathbf{u^{\ell} u^{\ell}}} (\bm{\mu}^{(\ell)}- \mathbf{m}_{\ubf^{\ell}} ),
    \end{split}
    \\
    \begin{split}
    \widetilde{\mathbf{S}}_{\fbf}^{(\ell)} &\defines \widetilde{\mathbf{S}} (\fbf^{(\ell)}, \fbf^{(\ell)})
    \\
    &=
    \mathbf{K_{f^{\ell} f^{\ell}}} - \mathbf{K_{f^{\ell} u^{\ell}}}\mathbf{K}_{\mathbf{u^{\ell} u^{\ell}}} (\mathbf{K}_{\mathbf{u^{\ell} u^{\ell}}} - \bm{\Sigma}^{(\ell)}) \mathbf{K}_{\mathbf{u^{\ell} u^{\ell}}}\mathbf{K_{u^{\ell} f^{\ell}}},
    \end{split}
\end{align}
with mean functions $\mathbf{m_{\fbf^{\ell}}} \defines m(\fbf^{(\ell - 1)})$ and $\mathbf{m_{\ubf^{\ell}}} \defines m(\Zbf^{(\ell - 1)})$ and inducing inputs $\Zbf^{(\ell - 1)}$ for $\ell = 1,...,L$.

The marginals within each layer thus only depend on the corresponding inputs, and so the $n$th marginal of the final layer of the \dgp posterior predictive distribution can be expressed as
\begin{align}
\label{eq:gpposterior}
    q(\fbf_n^{(L)})
    =
    \int \prod_{\ell =1}^{L-1} q(\fbf_n^{(\ell)} | \bm{\mu}^{(\ell)}, \bm{\Sigma}^{(\ell)}; \fbf_n^{(\ell - 1)}, \Zbf^{(\ell - 1)}) \dfbf_n^{(\ell)},
\end{align}
where $\fbf_n^{(\ell)}$ is the $n$th row of $\fbf^{(\ell)}$. This quantity is easy to compute using the reparameterization trick, that allows for sampling from the $n$th instances of the variational posteriors across layers by defining
\begin{align}
\label{appeq:reparameterizationtrick}
    \hat{\fbf}_n^{(\ell)}
    \defines
    \widetilde{\mathbf{m}}\left(\hat{\fbf}_n^{(\ell - 1)}\right) + \bm{\epsilon}_n^{(\ell)} \odot \sqrt{\widetilde{\mathbf{S}} \left(\hat{\fbf}_n^{(\ell - 1)}, \hat{\fbf}_n^{(\ell - 1)}\right) } 
\end{align}
and sampling from $\bm{\epsilon}_n^{(\ell)} \sim \mathcal{N}(\mathbf{0},\mathbf{I}_{D^{(\ell)}})$ \citep{Kingma2014,Salimbeni2017}.


\section{Derivation of Evidence Lower Bound}
\label{appsec:elbo}

Starting from the log-likelihood,
\begin{align}
    \log p(\ybf)
    =
    \log \mathbb { E } _ { q ( \{ \mathbf { f } ^ { (\ell) } , \mathbf { u } ^ { (\ell) } \} _ { \ell = 1 } ^ { L } ) } \left( \frac { p ( \mathbf { y } , \{ \mathbf { f } ^ { (\ell) } , \mathbf { u } ^ { (\ell) } \} _ { \ell = 1 } ^ { L } ) } { q ( \{ \mathbf { f } ^ { (\ell) } , \mathbf { u } ^ { (\ell) } \} _ { \ell = 1 } ^ { L } ) } \right),
\end{align}
with variational posterior
\begin{align}
    \mathcal{Q}
    =
    q(\{ \fbf^{\ell} , \ubf^{(\ell)} \}_{\ell = 1}^L ) = \prod_{\ell = 1}^{L} p(\fbf^{(\ell)} | \ubf^{(\ell)}, \fbf^{(\ell - 1)}) q(\ubf^{(\ell)})
\end{align}
and joint distribution
\begin{align}
    p(\ybf, \{ \fbf^{(\ell)}, \ubf^{(\ell)} \}_{\ell = 1}^ {L} )
    =
    \prod_{n=1}^N p(\ybf_n | \fbf_n^{(L)}) \prod_{\ell =1}^{L} p(\fbf^{(\ell)} | \ubf^{(\ell)}; \fbf^{(\ell - 1)}, \bm{\Omega}^{(\ell - 1)}) p(\ubf^{(\ell)} ; \bm{\Omega}^{(\ell - 1)}),
\end{align}
Lower bounding it by applying Jensen's inequality, we get the evidence lower bound
\begin{align}
\begin{split}
    \log p(\ybf)
    &=
    \log \mathbb { E } _ { q ( \{ \mathbf { f } ^ { (\ell) } , \mathbf { u } ^ { (\ell) } \} _ { \ell = 1 } ^ { L } ) } \left[ \frac { p ( \mathbf { y } , \{ \mathbf { f } ^ { (\ell) } , \mathbf { u } ^ { (\ell) } \} _ { \ell = 1 } ^ { L } ) } { q ( \{ \mathbf { f } ^ { (\ell) } , \mathbf { u } ^ { (\ell) } \} _ { \ell = 1 } ^ { L } ) } \right]
    \\
    &\geq
    \mathbb { E } _ { q ( \{ \mathbf { f } ^ { (\ell) } , \mathbf { u } ^ { (\ell) } \} _ { \ell = 1 } ^ { L } ) } \left[ \log \left( \frac { p ( \mathbf { y } , \{ \mathbf { f } ^ { (\ell) } , \mathbf { u } ^ { (\ell) } \} _ { \ell = 1 } ^ { L } ) } { q ( \{ \mathbf { f } ^ { (\ell) } , \mathbf { u } ^ { (\ell) } \} _ { \ell = 1 } ^ { L } ) } \right) \right] \defines \calL.
\end{split}
\end{align}
Writing the expectation as an integral and substituting in the variational posterior and joint distribution, we get
\begin{align}
\begin{split}
    \calL &= \int \int q ( \{ \mathbf { f } ^ { (\ell) } , \mathbf { u } ^ { (\ell) } \} _ { \ell = 1 } ^ { L } ) \log \left( \frac { p ( \mathbf { y } , \{ \mathbf { f } ^ { \ell } , \mathbf { u } ^ { (\ell) } \} _ { \ell = 1 } ^ { L } ) } { q ( \{ \mathbf { f } ^ { (\ell) } , \mathbf { u } ^ { (\ell) } \} _ { \ell = 1 } ^ { L } ) } \right) \dee \{\fbf^{(\ell)}, \ubf^{(\ell)} \}^{L}_{\ell=1}
    \\
    &=
    \int \int q ( \{ \mathbf { f } ^ { (\ell) } , \mathbf { u } ^ { (\ell) } \} _ { \ell = 1 } ^ { L } ) \log \left( \frac { \prod _ { n = 1 } ^ { N } p \left( \mathbf { y } _ { n } | \mathbf { f } _ { i } ^ { L } \right) \prod _ { l = 1 } ^ { L } p \left( \mathbf { f } ^ { \ell } | \mathbf { u } ^ { \ell } ; \mathbf { f } ^ { l - 1 } , \bm { \Omega } ^ { l - 1 } \right) p \left( \mathbf { u } ^ { \ell } ; \bm { \Omega } ^ { l - 1 } \right) } { \prod _ { l = 1 } ^ { L } p \left( \mathbf { f } ^ { \ell } | \mathbf { u } ^ { \ell } ; \mathbf { f } ^ { l - 1 } ; \bm { \Omega } ^ { l - 1 } \right) q \left( \mathbf { u } ^ { \ell } \right) } \right) \dee \{\fbf^{(\ell)}, \ubf^{(\ell)} \}^{L}_{\ell=1}.
\end{split}
\end{align}
Cancelling out the identical terms in the logarithm and rewriting the resulting expression, we get
\begin{align}
\begin{split}
    \calL
    &=
    \iint q ( \{ \mathbf { f } ^ { (\ell) } , \mathbf { u } ^ { (\ell) } \} _ { \ell = 1 } ^ { L } ) \log \left( \frac { \prod _ { n = 1 } ^ { N } p \left( \mathbf { y } _ { n } | \mathbf { f } _ { n } ^ { L } \right) \prod _ { l = 1 } ^ { L } p( \mathbf { u } ^ { \ell } ; \bm { \Omega } ^ { l - 1 }) } { \prod _ { l = 1 } ^ { L } q \left( \mathbf { u } ^ { \ell } \right) } \right) \dee \{\fbf^{(\ell)}, \ubf^{(\ell)} \}^{L}_{\ell=1}
    \\
    &=
    \iint q ( \{ \mathbf { f } ^ { (\ell) } , \mathbf { u } ^ { (\ell) } \} _ { \ell = 1 } ^ { L } ) \log \prod _ { n = 1 } ^ { N } p( \mathbf { y } _ { n } | \mathbf { f } _ { n } ^ { L } )\, d\{\fbf^{(\ell)}, \ubf^{(\ell)} \}^{L}_{\ell=1}
    \\
    &\qquad
    + \iint q ( \{ \mathbf { f } ^ { (\ell) } , \mathbf { u } ^ { (\ell) } \} _ { \ell = 1 } ^ { L } ) \log \left( \frac { \prod _ { l = 1 } ^ { L } p \left( \mathbf { u } ^ { \ell } ; \bm{\Omega}^{(\ell - 1)} \right) } { \prod _ { l = 1 } ^ { L } q \left( \mathbf { u } ^ { \ell } \right) } \right) \dee \{\fbf^{(\ell)}, \ubf^{(\ell)} \}^{L}_{\ell=1}
    \\
    &=
    \int q ( \{ \mathbf { f } ^ { (\ell) } , \mathbf { u } ^ { (\ell) } \} _ { \ell = 1 } ^ { L } ) \log \prod _ { n = 1 } ^ { N } p( \mathbf { y } _ { n } | \mathbf { f } _ { n } ^ { L } ) \dee \{ \mathbf { f } ^ { \ell } \} _ { l = 1 } ^ { L }
    \\
    &\qquad
    + \iint q ( \mathbf { u } ^ { (\ell) } \} _ { \ell = 1 } ^ { L } ) \log \left( \frac { \prod _ { l = 1 } ^ { L } p \left( \mathbf { u } ^ { \ell } ; \bm{\Omega}^{(\ell - 1)} \right) } { \prod _ { l = 1 } ^ { L } q \left( \mathbf { u } ^ { \ell } \right) } \right) \dee \{\ubf^{(\ell)} \}^{L}_{\ell=1}
    \\
    &=
    \sum_{n=1}^N \E_{q(\fbf_n^{(L)})} \big[ \log p(\ybf_n | \fbf_n^{(L)}) \big] - \sum_{\ell = 1}^{L} \text{KL}(q(\ubf^{(\ell)}) \, \vert \vert \,  p(\ubf^{(\ell)})).
\end{split}
\end{align}

\end{appendices}

\end{document}